\documentclass[runningheads]{llncs}

\usepackage[mobile]{accv}

\usepackage{accvabbrv}

\usepackage{graphicx}
\usepackage{booktabs}

\usepackage{algorithm}
\usepackage{xcolor}
\usepackage{pifont}
\usepackage{amssymb}
\usepackage{multirow}
\usepackage{overpic}
\usepackage{colortbl}
\usepackage{lipsum}
\usepackage{mdframed}
\usepackage{soul}
\usepackage{tikz}
\usetikzlibrary{spy, positioning}
\usepackage[normalem]{ulem}

\usepackage[accsupp]{axessibility}  %

\usepackage[breaklinks,colorlinks,citecolor=accvblue]{hyperref}

\usepackage{orcidlink}

\newcommand{\hlc}[2][yellow]{{%
    \colorlet{foo}{#1}%
    \sethlcolor{foo}\hl{#2}}%
}
\newcommand{\best}{\cellcolor{red!30}}
\newcommand{\sbest}{\cellcolor{orange!30}}
\newcommand{\tbest}{\cellcolor{yellow!30}}

\begin{document}

\newcommand{\method}{Cyclone}
\title{
\method: Diffusion Model for Cycle-Consistent Weather Editing from Unpaired Driving Data}

\titlerunning{\method}

\author{Thang-Anh-Quan Nguyen\inst{1,2}\orcidlink{0009-0004-4001-4279} \and
Moussab Bennehar\inst{1}\orcidlink{0000-0002-6566-6132} \and 
Luis Rold{\~a}o\inst{1}\orcidlink{0000-0003-0482-3584} \and
\\Nathan Piasco\inst{1}\orcidlink{0000-0001-7952-6643} \and 
Dzmitry Tsishkou\inst{1}\orcidlink{0009-0002-9798-3316} \and
Laurent Caraffa\inst{3}\orcidlink{0000-0002-8676-8058} \and
\\Jean-Philippe Tarel\inst{2}\orcidlink{0000-0002-9241-5347} \and
Roland Br{\'e}mond\inst{2}\orcidlink{0000-0003-3150-7624}
}

\authorrunning{T.A.Q.~Nguyen et al.}

\institute{Noah's Ark, Huawei Paris Research Center, France \and
COSYS, Gustave Eiffel University, France \and
LASTIG, IGN-ENSG, Gustave Eiffel University, France}

\maketitle

\begin{figure}
  \vspace{-0.5cm}
  \centering
  \includegraphics[width=0.9\linewidth]{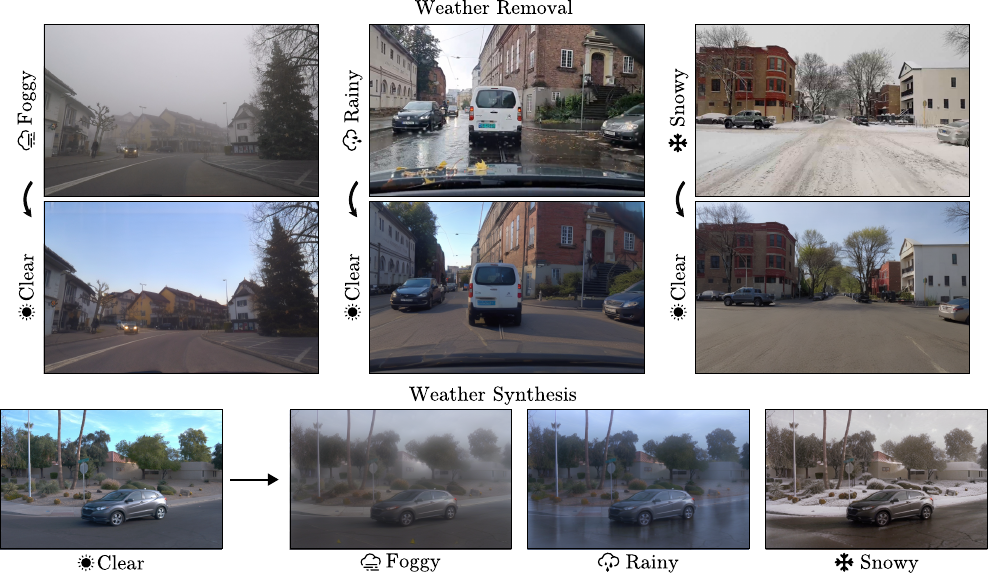}
  \caption{\textbf{All-in-One Realistic Weather Editing.} We introduce \method, a diffusion-based framework for weather editing in autonomous driving. It can both synthesize high-fidelity diverse weather conditions (bottom) and remove adverse effects (top). Notably, this is achieved without requiring any paired data during training.}
  \label{fig:teaser}
  \vspace{-1cm}
\end{figure}

\begin{abstract}
Reliable perception under diverse weather conditions remains a major challenge for autonomous driving systems. A common strategy to improve robustness is either to synthesize adverse weather conditions for training perception models or to apply weather-removal techniques to recover clean inputs. However, existing approaches typically rely on synthetic data augmentation or physics-based, task-specific models that require paired training data and often struggle to generate realistic weather effects or generalize robustly to out-of-domain scenarios.
Toward this problem, we present Cyclone, a unified framework for weather editing based on latent diffusion, equipped with cycle-consistent constraints and knowledge from image-text models. Cyclone enables the generation of multiple weather conditions across diverse scenes while eliminating the need for paired data. Experimental results show that our approach produces more realistic, structure-preserving outputs than existing baselines and leads to consistent improvements across several downstream driving perception tasks. Furthermore, we demonstrate that Cyclone can be distilled to a video diffusion model for temporally consistent weather editing. 
\end{abstract}
    
\section{Introduction}
\label{sec:intro}

Weather synthesis and removal are fundamental tasks in autonomous driving and mobile robotics, where adverse environmental conditions such as rain, fog, and snow can drastically degrade sensor accuracy and, consequently, perception performance. Achieving robust perception across such diverse conditions is therefore essential for safe and reliable navigation. Without modifying sensor hardware, a widely adopted strategy involves training and evaluating perception models~\cite{sakaridis2021acdc, sun2022shift} under adverse weather conditions either through synthetic weather generation or applying weather-removal techniques on degraded inputs.

Current state-of-the-art methods typically tackle weather synthesis and removal as an image-to-image (I2I) translation~\cite{isola2017image} problem. However, constructing high-quality paired datasets from the same scene under different weather conditions to supervise such models is practically infeasible due to real-world continuous dynamics. 
This inherent challenge in the data acquisition step has driven most existing methods toward using physics-based modeling and/or synthetic data augmentation. While these approaches can simulate specific weather effects, their hand-crafted physical priors provide only simplified approximations of real-world phenomena and are not broadly applicable across different weather conditions. As a result, their performance often degrades in real-world scenarios due to domain gaps and the complexity of natural weather.
Motivated by the strong representational capacity of modern generative models, some works~\cite{meng2021sdedit, hertz2022prompt} explore inverting diffusion models trained on large-scale data. Nevertheless, they primarily operate at test time and target object- or region-level edits rather than faithfully capturing the complexity and spatial coherence of the weather effects as a whole.
Overall, both often fall short in terms of structural fidelity and visual realism when applied to practical scenarios.

Our work aims to improve weather synthesis from driving data for diffusion models \textit{without training pairs}.
Inspired by the cycle-consistency used in CycleGAN~\cite{zhu2017unpaired}, we introduce a constraint that encourages forward and backward translations to be mutually consistent, thereby preserving scene geometry and preventing structural drift in the translated outputs.
However, unlike CycleGAN, we eliminate the need for multiple generators and discriminators for every domain by leveraging pretrained diffusion models. This is made possible by two key factors: (1) diffusion models are large-scale image generators~\cite{dhariwal2021diffusion} that can naturally support multi-condition image generation within a single network~\cite{brooks2023instructpix2pix}, and (2) models trained with text descriptions capture rich high-level semantics. Such powerful pretrained models can be employed to assess whether the generated/rendered image is realistic or not, akin to the discriminator in GANs~\cite{poole2022dreamfusion, mukhopadhyay2023diffusion, sauer2024fast}.
Therefore, our approach can naturally scale to multiple weather conditions without collapsing. Moreover, we find that the cycle constraint alone is insufficient, as the model can easily ``overfit" to the input condition. We further propose using knowledge distillation from pretrained text-to-image models, which acts as a teacher to preserve strong compositional and semantic priors learned from large-scale data. This guidance helps maintain prior domain knowledge while improving translation faithfulness and visual quality.

Finally, we demonstrate several practical applications of our model. Cyclone enhances downstream perception tasks (\eg, depth estimation, semantic segmentation, and object detection) under adverse weather conditions, highlighting its potential for real-world deployment in autonomous driving and robotics. Cyclone can serve as an auxiliary data source for finetuning other models, such as video diffusion, which is capable of performing consistent weather editing given video inputs.
Our contributions can be summarized as follows:
\begin{itemize}
    \item We propose a novel cycle-consistent training strategy for diffusion model, leveraging its strong generative representation for weather editing tasks from unpaired data.
    \item We introduce a distillation technique that provides guidance and regularizes training, preventing overfitting.
    \item Extensive experiments that demonstrate the strong performance in weather editing compared to SoTA, which heavily relied on paired data.
    \item Autonomous driving tailored applications, such as distilling a video diffusion model for temporally consistent video editing and improving downstream perception tasks.
\end{itemize}

\section{Related Work}
\label{sec:related_work}

\noindent\textbf{Image and Video Editing} 
are fundamental tasks in generative models, with the goal of altering content or style while preserving scene structure and semantics.
While early efforts, such as CycleGAN~\cite{zhu2017unpaired} and its variants~\cite{yi2017dualgan, liu2017unsupervised, torbunov2023uvcgan} are \textit{task-agnostic} and capable of learning from unpaired data, they typically require training a separate generator and discriminator for each domain, which incurs high computational costs and limits scalability as well as flexibility when handling diverse or evolving domains.
On the other hand, diffusion-based models offer a more expressive and scalable alternative by directly modeling data distributions through iterative denoising. This allows approaches such as SDEdit~\cite{meng2021sdedit} and DDIM Inversion~\cite{song2020denoising} to perform editing at inference time and to generalize across multiple text prompts. InstructPix2Pix~\cite{brooks2023instructpix2pix} introduces instruction-based image editing by fine-tuning a diffusion model on data produced by Prompt-to-Prompt~\cite{hertz2022prompt}.  OmniGen~\cite{xiao2025omnigen}, BAGEL~\cite{deng2025emerging}, Flux-Kontext~\cite{labs2025flux}, and Qwen-Image-Edit~\cite{bai2025qwen2} all follow a similar paradigm by scaling model capacity and relying on labor-intensive data curation pipelines for general-purpose image editing. However, these models depend heavily on paired or curated supervision and are primarily designed for broad editing tasks rather than safety-critical applications. As a result, they are often heavily-built and less suitable for domain-specific settings due to data reliability.

Bringing cycle-consistency to diffusion models, CycleNet~\cite{xu2023cyclenet}, Img2Img-Turbo~\cite{parmar2024one} represent the closest prior work to ours. While they also target unpaired training with diffusion models, their architectural and training designs also lack sufficient regularization to prevent overfitting and to ensure robustness at scale, particularly when handling multiple domains or large datasets. NP-Edit~\cite{kumari2025learning} relies on a VLM critic during training and therefore lacks pixel-wise supervision. As a result, the edited images may deviate from the input in spatial alignment.
Furthermore, most prior work is designed and evaluated for aesthetic editing, mostly on object-centric data (e.g., horse-to-zebra or cat-to-dog, object removal/inpainting). Consequently, these methods often struggle to perform substantial modifications on scene-level complexity.

\noindent\textbf{Weather Synthesis and Removal.}
This topic has been a long-standing focus in computer vision, traditionally treated as a \textit{task-specific} problem targeting applications such as de-hazing~\cite{wu2023ridcp, lan2025exploiting}, de-raining~\cite{yang2017deep}, de-snowing~\cite{chen2020jstasr}.  
Classical approaches often rely on physically grounded models to characterize the degradation process.  For instance, the Koschmieder's law or Atmospheric Scattering Model~\cite{nayar1999vision} describes hazy images as: $\mathbf{I} = \mathbf{J}e^{-\beta d}+\mathbf{A}(1 - e^{-\beta d})$, where $\mathbf{J}$ denotes the image without haze, $\beta$ is the scattering coefficient, $d$ represents the seen object depth, and $\mathbf{A}$ is the global atmospheric light. Similarly, rain and snow are commonly modeled as additive transients~\cite{qian2018attentive, liu2018desnownet,li2019heavy}, often expressed as: $\mathbf{I} = \mathbf{J}\odot(1 - \mathbf{M}) + \mathbf{T}$, where $\mathbf{M}$ is a mask indicating weather-affected regions and $\mathbf{T}$ captures rain streaks or snowflakes.
More recent trends in the field show a shift toward more flexible architectures capable of handling multiple weather conditions simultaneously~\cite{ozdenizci2023restoring, ye2023adverse, valanarasu2022transweather, sun2024restoring, rajagopalan2025awracle}, where learnable weather-type embeddings are used to dynamically adapt to specific degradation.
These methods face significant limitations in practical deployment because they typically require paired datasets to supervise these restoration processes.
However, collecting such paired real-world data under diverse and dynamic weather conditions is extremely challenging, if not infeasible. This reliance severely limits diversity and scalability, thus slowing community advancements by tying advances to the availability of new datasets or simulation engines~\cite{lin2025controllable, zhu2025weatherdiffusion}.
Therefore, we propose a task-agnostic diffusion-based framework that can both add and remove multiple weather conditions within a single model, without requiring paired data during training.

\noindent\textbf{Driving Diffusion Models.}
Recent advances in generative AI have demonstrated that world models for autonomous driving can be trained on large-scale data for various downstream applications.  However, existing architectures are typically designed as image-to-video and text-to-video models. Image-to-video models~\cite{yang2024generalized, wang2024driving, wang2024drivedreamer, gao2024vista} focus on next-frame generation or view-synthesis, emphasizing spatial-temporal consistency and scene dynamics. In contrast, text-to-video models~\cite{gao2023magicdrive, ren2025cosmos} demonstrate the ability to synthesize diverse scenes with controllable appearance through textual input and geometric conditioning. Although both types of models perform well on several generative tasks, neither is designed for scene-consistent weather editing. In practice, they tend to either under-stylize the weather effects, failing to produce noticeable changes, or over-stylize the scene, disrupting the original structure and appearance.

\section{Preliminaries}

\subsection{Diffusion Models}
Diffusion models~\cite{ho2020denoising} are probabilistic generative models that aim to capture the underlying data distribution $p(x)$ from a noise distribution. They operate by iteratively denoising a Gaussian variable $x_{T}$ to reconstruct the original data sample $x_{0}$. In Latent Diffusion Models (LDMs)~\cite{rombach2022high}, this diffusion process is rather performed in a compressed latent space obtained via a Variational Autoencoder (VAE)~\cite{kingma2013auto} to reduce computational complexity.
The denoising network $\epsilon_{\phi}$, typically a U-Net~\cite{ronneberger2015u}, is trained to predict noise conditioned on input signals by minimizing:
\begin{equation}
    \mathcal{L}(\phi)=\mathbb{E}_{t, \epsilon}\|\epsilon_{\phi}(x_t,c)-\epsilon\|^{2}_{2},
\label{e:eq1}
\end{equation}
where $\epsilon\sim \mathcal{N}(0, I)$ is the additive Gaussian noise, $t\sim\mathcal{U}(0, T)$ is the time step, $x_t$ is the noisy latent at $t$, and $c$ denotes the condition signal, which is used for conditional generation.

\subsection{CycleGAN}
The goal of unpaired I2I translation is to learn a mapping between two domains using unpaired training samples. 
Specifically, given two sets of samples $\{x^i\}_{i=1}^N$ with $x^i \in \mathcal{X}$ labeled with $c_x$, and $\{y^j\}_{j=1}^M$ with $y^j \in \mathcal{Y}$ labeled with $c_y$ without one-to-one correspondences, \ie $i\neq j$, the objective is to learn transformations $x^i\rightarrow y^i$ or $y^j\rightarrow x^j$ while preserving the input content. 
CycleGAN~\cite{zhu2017unpaired} tackles this problem by combining adversarial losses, which encourage generated outputs to be indistinguishable from real samples in the target domain, with a cycle-consistency constraint, which enforces structural preservation by reconstructing the original image when translating back to the source domain. 
We extend this principle to LDMs while addressing key limitations of prior cycle-consistent methods, including the mode collapse and low generalizability observed in CycleGAN~\cite{zhu2017unpaired} and the identity overfitting exhibited by CycleNet~\cite{xu2023cyclenet}. Furthermore, rather than training separate networks for each domain, our approach employs a single unified denoising model that flexibly handles translation across multiple weather conditions.

\section{Method}
\label{sec:method}

\begin{figure*}
\centering    
\includegraphics[width=\linewidth]{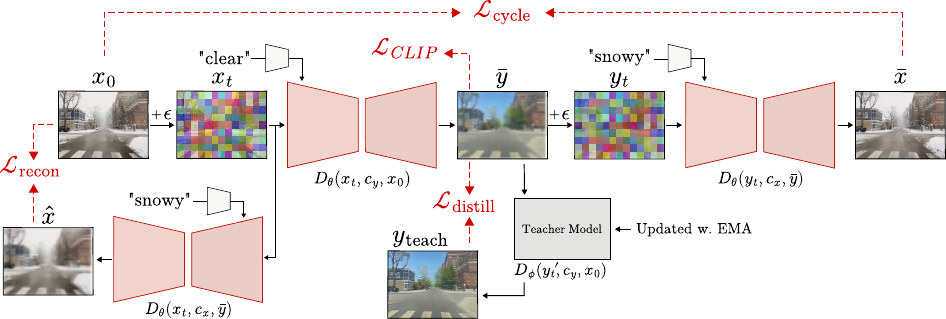}
\caption{\textbf{Overview of Cyclone's Training Pipeline.} 
Our image translation pipeline includes a forward translation from $x_t$ to $\bar{y}$, conditioned on the text prompt $c_y$ and input image $x_0$, followed by a backward translation from $y_t$ to $\bar{x}$, conditioned on $c_x$ and $\bar{y}$. We enforce this cycle consistency using $\mathcal{L}_{\text{recon}}$ and $\mathcal{L}_{\text{cycle}}$. To further stabilize training and prevent collapse to the trivial identity mapping, we introduce a teacher model $D_\phi$ that provides regularization. The teacher is initialized with the same weights as the student model and is subsequently updated using an EMA of the student parameters. In particular, we use the same student model $D_\theta$ across multiple domains. \textit{Concatenation of the conditioning image is omitted for simplicity.}}
\label{fig:overview}
\end{figure*}

\subsection{Problem Statement}
In the following, we focus only on the translation from an arbitrary weather domain $\mathcal{X}$ to $\mathcal{Y}$, given the symmetry of the backward translation. We also treat $x$ and $y$ interchangeably as image or latent representations, since they can be encoded and decoded using a VAE.
Our goal is to learn a denoiser $D_{\theta}$ capable of performing weather transitions from an original scene image $x_0$ and a text prompt $c_y$ describing the target weather. We consider our set of weather effects as ``clear'', ``foggy'', ``rainy'', and ``snowy''. During inference, as the model iteratively denoises a noisy latent $y_T$ to $y_0$, it produces an adaptation of the source image under the specified prompt. We denote the output as $D_{\theta}(y_t, c_y, x_0)$ and modify the LDM's input layer to accept both noisy input $y_t$ and clean source image $x_0$ using channel-wise concatenation while injecting the textual input $c_y$ through CLIP encoder~\cite{radford2021learning} and cross-attention layers. To achieve this output, we present the training pipeline in~\cref{fig:overview} and in the subsections below.

\subsection{Cycle-Consistency}
We define two objectives to enforce cycle consistency. In particular, the reconstruction loss $\mathcal{L}_{\text{recon}}$ ensures that the model can reverse a noisy input $x_t$ similar to~\cref{e:eq1}.
The cycle loss $\mathcal{L}_{\text{cycle}}$ enforces cross-domain consistency under a forward $x_0\rightarrow \bar{y}=D_{\theta}(x_t, c_y, x_0)$ and a backward $\bar{y} \rightarrow \bar{x}=D_{\theta}(y_t, c_x, \bar{y})$ translation by requiring the model to recover the input $x_0$ when denoising the translated latent $y_t$, given the source condition $c_x$ and the target condition $c_y$:
\begin{equation}
    \mathcal{L}_{\text{recon}} = \mathbb{E}_{t}\|\hat{x} - x_0 \|^2_2=\mathbb{E}_{t}\|D_{\theta}(x_t, c_x, \bar{y})-x_0\|^2_2,
    \label{eq:loss_recon}
\end{equation}

\begin{equation}
    \mathcal{L}_{\text{cycle}} = \mathbb{E}_{t}\|\bar{x} - x_0 \|^2_2=\mathbb{E}_{t}\|D_{\theta}(y_t, c_x, \bar{y}) - x_0\|^2_2.
    \label{eq:loss_cycle}
\end{equation}

\noindent\textbf{Invariance  via Multi-Domain Consistency.}
The training is not limited to pairwise domains but instead encompasses multiple weather conditions, thus reinforcing translation invariance. To further analyze this, we introduce $\mathcal{Z}$ as an arbitrary alternative domain distinct from $\mathcal{Y}$ and apply the same losses in~\cref{eq:loss_recon} and~\cref{eq:loss_cycle}. Intuitively, by conditioning on the same target text, different weather images of the same scene lead to similar enough predictions, so that the target image stays invariant. Therefore, one must ensure that when conditioning from multiple domains $(\mathcal{Y}, \mathcal{Z},...)$ back to the source $\mathcal{X}$, all reconstruct the same content. Applying the triangle inequality followed by Jensen's inequality yields a timestep-dependent bound:
\begin{equation}
    \mathbb{E}_{t}\|D_{\theta}(x_t, c_x, \bar{y}) - D_{\theta}(x_t, c_x, \bar{z})\| \le \sqrt{\mathcal{L}_{\text{recon}}} + \sqrt{\mathcal{L}'_{\text{recon}}}\triangleq B_t,
\end{equation}
where $\mathcal{L}'_{\text{recon}} =\mathbb{E}_{t}\|D_{\theta}(x_t, c_x, \bar{z}) - x_0\|^2_2$. Formally, upon convergence, the objectives are minimized, bounding the expected reconstruction errors and resulting in a small $B_t$. Invariance emerges implicitly from geometric constraints imposed by multi-domain cycle consistency, without the need to introduce any constraint, unlike the invariance loss used in CycleNet~\cite{xu2023cyclenet}.

\subsection{Diffusion Prior Distillation}
Since the ground-truth target image $y_0$ is unavailable, the two losses can be trivially minimized by ignoring the text condition $c_y$ and simply outputting identity (\ie, the condition image $x_0$ itself). This issue does not arise in CycleGAN~\cite{zhu2017unpaired}, as each model is trained for a specific domain pair with its own discriminator. However, extending this design to multiple weather domains would require maintaining multiple discriminators, one per domain, which quickly becomes impractical. 
As diffusion models can inherently perform zero-shot classification~\cite{li2023your} and provide meaningful information from their learned data distribution, we draw inspiration from Score Distillation Sampling (SDS)~\cite{poole2022dreamfusion}. SDS is a distillation technique that leverages the gradient of a frozen diffusion model $\phi$ to guide another model $\theta$ (\eg, a 3D rendering model) to align with a given text condition:
\begin{equation}
    \nabla_{\theta}\mathcal{L}_{\text{SDS}}(x) \triangleq \mathbb{E}_{t, \epsilon} \left[ 
        w(t) \left(\epsilon_{\phi} - \epsilon \right)\frac{\partial x}{\partial \theta} 
    \right],
\end{equation}
where $w(t)$ is a weighting function at timestep $t$.
By omitting the U-Net Jacobian term in the gradient, the differentiable image generator $\theta$ can be optimized without back-propagating through the model’s U-Net, significantly reducing memory and computational costs compared to any additional discriminator.

A teacher network $\phi$ is introduced to provide a stable reference. It is initialized with the same weights as the student and is updated via an exponential moving average (EMA)~\cite{tarvainen2017mean} of the student’s parameters each training step, with a momentum of 0.9999. This approach encourages the student to produce outputs that are consistent with the teacher while allowing gradual refinement from cycle consistency losses, effectively acting as a regularizer that prevents collapse or identity mapping during training. The revised version of SDS loss is defined as:
\vspace{-0.2cm}
\begin{align}
    \mathcal{L}_{\text{distill}}
    &= \mathbb{E}_{t}\!\left[ 
        w(t)\left(\bar{y} - y_{\text{teach}}\right)
    \right] \nonumber \\
    &= \mathbb{E}_{t}\!\left[ 
        w(t)\left(D_{\theta}(x_t, c_y, x_0) - D_{\phi}(y'_{t}, c_y, x_0)\right)
    \right],
\end{align}
where $t'\sim\mathcal{U}(0, T)$. As a result, this distillation loss encourages the edited image to align more faithfully to the distribution of the real images modeled by the pretrained teacher.

\subsection{Textual Guidance}
Textual features have been shown to provide enriched high-level semantics and have demonstrated strong effectiveness in training diffusion models. Motivated by this observation, we further incorporate textual information to enhance the editing process through semantic guidance. Specifically, we aim to identify a weather transition direction in the CLIP embedding~\cite{radford2021learning} space that characterizes the shift between the source and target domains: 
\begin{equation}
    \mathcal{L}_{\text{CLIP}} = 1 - \frac{\langle\Delta I,\Delta T\rangle}{\|\Delta I\| \|\Delta T\|},
\end{equation}
$\Delta I=E_I(\bar{y})-E_I(x_0),~\Delta T=E_T(c_y)-E_T(c_x)$ where $E_I$ and $E_T$ are CLIP’s image and text encoders, respectively.
By jointly leveraging positive and negative samples, the CLIP text encoder mitigates the influence of mode collapse~\cite{gal2022stylegan}. The model is required to satisfy CLIP-based constraints across a diverse set of samples rather than overfitting to specific instances, thereby promoting more robust weather control.
The complete student training objective is as follows:
\begin{equation}
    \mathcal{L}_{\text{total}} = \mathcal{L}_{\text{recon}} + \lambda_{\text{cycle}}\mathcal{L}_{\text{cycle}} + \lambda_{\text{distill}}\mathcal{L}_{\text{distill}} + \lambda_{\text{CLIP}}\mathcal{L}_{\text{CLIP}},
\end{equation}
where $\lambda_{\text{cycle}}, \lambda_{\text{distill}}, \lambda_{\text{CLIP}}$  are weighting factors.
Finally, we adapt these formulations to align with the v-prediction parameterization.

\section{Experiments}
\label{sec:experiment}

\subsection{Experimental Setup}
\noindent\textbf{Implementation Details.}
We use Stable Diffusion v2.1~\cite{rombach2022high} as the base model. We only optimize the U-Net while keeping the VAE and text encoder frozen. 
Following Lin \etal~\cite{lin2024common}, we rescale the noise scheduler to enforce a zero terminal SNR and switch the sampling type to \texttt{trailing}. This modification prevents the model from fitting to the low-frequency components during training, which often correspond to existing weather or lighting conditions.
We set $\lambda_{\text{cycle}}=\lambda_{\text{distill}}=0.5, \lambda_{\text{CLIP}}=0.1$, and apply $\mathcal{L}_{\text{cycle}}$ when $t>0.5$, indicating that transitions should only occur under sufficiently high noise level~\cite{meng2021sdedit}.

We use AdamW~\cite{loshchilov2017decoupled} optimizer with a learning rate of 1e-5 and batch size of 8 for 100k training iterations. The training resolution is varied from $512\times 320$ to $1024\times 640$. During evaluation, we synthesize images at $512\times 320$ and subsequently resize them to match the input requirements of each evaluation model.

\noindent\textbf{Datasets.}
We train Cyclone on subsets of ACDC~\cite{sakaridis2021acdc}, SHIFT~\cite{sun2022shift}, BDD100K~\cite{yu2020bdd100k}, and OpenDV-YouTube~\cite{yang2024generalized} datasets. To ensure data quality, we manually removed irrelevant or corrupted content (\eg, poorly rendered frames or frames highly occluded by sun flare, windshield, and ego-vehicle), yielding approximately 2M images or around 50 hours of video at 10 FPS.
For evaluation, we use the validation sets of SHIFT~\cite{sun2022shift}, ACDC~\cite{sakaridis2021acdc}, alongside out-of-distribution datasets: nuScenes~\cite{caesar2020nuscenes}, PandaSet~\cite{xiao2021pandaset}, and Waymo~\cite{sun2020scalability}. Our curated validation set consists of 120 videos across 120 scenes, well-labeled with various types of weather. Additional details are available in the Supplementary Material.

\noindent\textbf{Metrics.}
Similar to prior work~\cite{parmar2024one, lin2025controllable}, we evaluate each generated image from multiple complementary metrics. 
For structural preservation, we adopt DINO-Struct ($\times 100$) based on DINOv2~\cite{oquab2023dinov2} to measure structural correspondence between the edited output and the input image.
To assess semantic alignment, we compute CLIP Score ($\times 100$)~\cite{hessel2021clipscore}, which quantifies the similarity between the edited images and their associated text prompts.
In addition, we employ CLIP~\cite{radford2021learning} as a weather classifier/text-retriever and report the resulting classification accuracy (\%).
Finally, we calculate FID~\cite{heusel2017gans} to evaluate how well the generated distribution matches the reference data sharing the same weather label. These metrics underscore the importance of jointly evaluating both distributional similarity and structural preservation, as favoring one at the expense of the other leads to suboptimal outcomes.

\noindent\textbf{Baselines.}
We divide the evaluation into two tasks: adding and removing weather, and compare our approach against several representative baselines.
Histoformer~\cite{sun2024restoring} and AWRaCLe~\cite{rajagopalan2025awracle} are \textit{specific weather restoration methods}, while InstructPix2Pix~\cite{brooks2023instructpix2pix}, BAGEL~\cite{deng2025emerging}, and Qwen-Image-Edit~\cite{bai2025qwen2} are SoTA in \textit{diffusion-based image editing}, all using pairs to train their model.
TokenFlow~\cite{geyer2023tokenflow} leverages textual instructions and propagates diffusion features based on inter-frame correspondences to achieve consistent video editing.
Finally, we evaluate against CycleGAN~\cite{zhu2017unpaired} and CycleNet~\cite{xu2023cyclenet}, models that incorporate \textit{cycle-consistency constraints}. We train TokenFlow, CycleGAN, and CycleNet using the same training data as Cyclone while using official pre-trained checkpoints for the remaining baselines due to their reliance on paired data.

\subsection{Quantitative Evaluation}
\cref{tab:quantitative} presents quantitative comparisons of our method against all baselines.
Histoformer and AWRaCLe achieve low DINO-Struct, but exhibit poor CLIP, classification accuracy, and FID, suggesting that they make only minimal changes to the input and therefore fail to perform effective weather removal.
InstructPix2Pix, BAGEL, and Qwen-Image-Edit produce highly unbalanced performance between the two tasks. 
Although CycleGAN and CycleNet adopt a cycle-consistency framework similar to ours, they exhibit clear signs of underfitting, as evidenced by their low CLIP and high DINO Scores.
On the other hand, our approach consistently achieves strong performance across all metrics, demonstrating its effectiveness in both generating and removing weather. Moreover, compared with methods trained on the same datasets, our approach reduces FID by approximately $5\sim10$, indicating a better ability to capture the target domain from unpaired data. Cyclone is also considerably more lightweight than large-scale models such as Qwen-Image-Edit, using 20× fewer parameters and achieving 100× faster inference.

\def\fgsize{0.16}

\begin{figure*}
\centering
\setlength{\tabcolsep}{0.002\linewidth}
\renewcommand{\arraystretch}{1.5}
\begin{tabular}{llcccccc}
    {} & {} & $+$Foggy & $+$Rainy & $+$Snowy & $-$Foggy & $-$Rainy & $-$Snowy\\ 
    \multirow{2}{*}[9mm]{\rotatebox[origin=c]{90}{Input}} & {} &
    \includegraphics[width=\fgsize\textwidth]{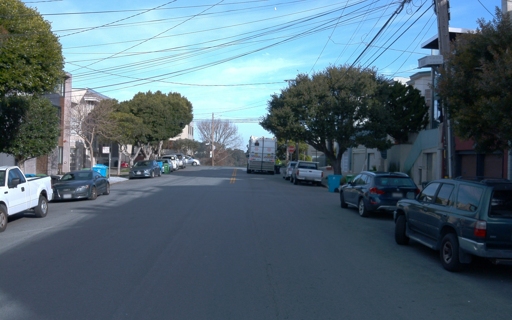} &
    \includegraphics[width=\fgsize\textwidth]{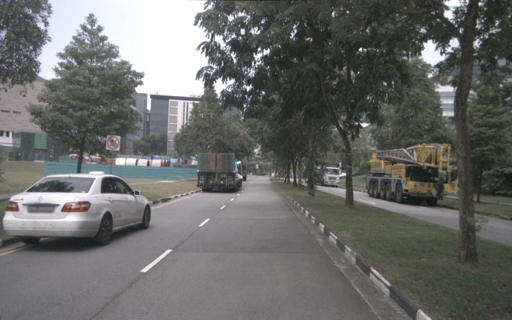} &
    \includegraphics[width=\fgsize\textwidth]{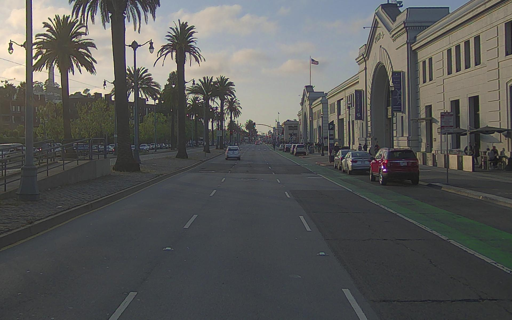} &
    \includegraphics[width=\fgsize\textwidth]{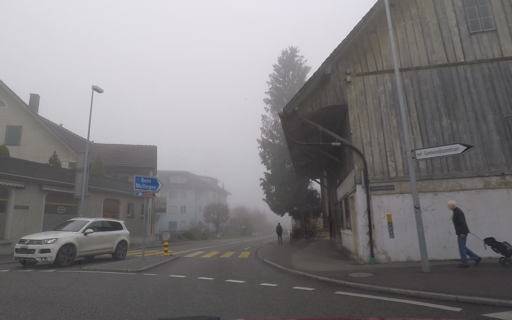} &
    \includegraphics[width=\fgsize\textwidth]{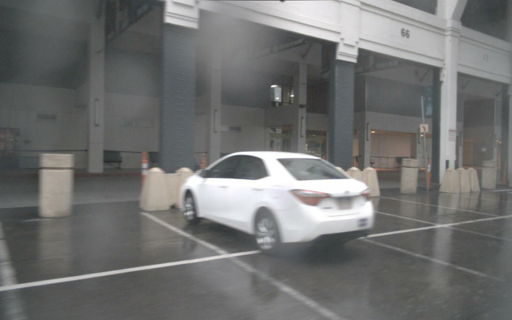} &
    \includegraphics[width=\fgsize\textwidth]{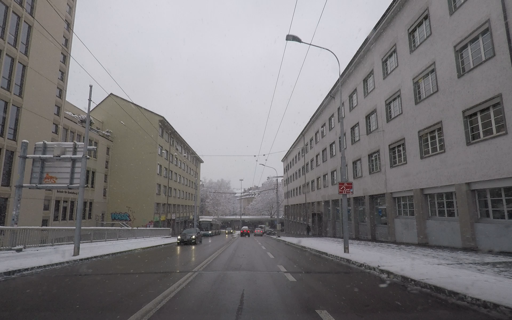}\\

    \multirow{2}{*}[9mm]{\rotatebox[origin=c]{90}{\small Histofor-}} & \multirow{2}{*}[9mm]{\rotatebox[origin=c]{90}{mer~\cite{sun2024restoring}}} &
    & & &
    \includegraphics[width=\fgsize\textwidth]{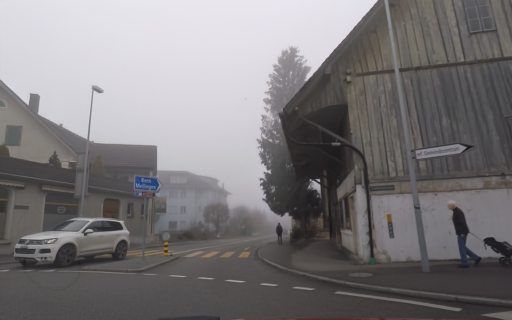} &
    \includegraphics[width=\fgsize\textwidth]{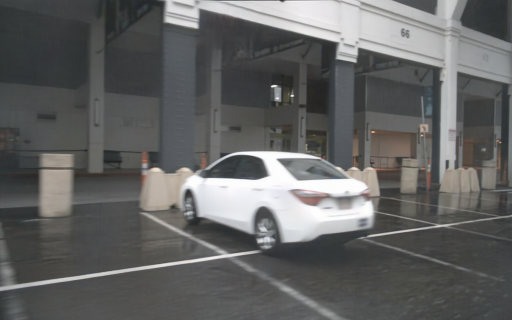} &
    \includegraphics[width=\fgsize\textwidth]{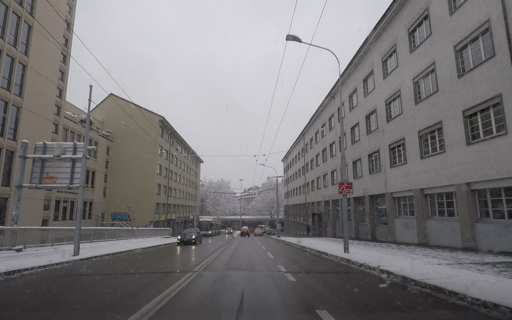}\\

    \multirow{2}{*}[9mm]{\rotatebox[origin=c]{90}{\small AwRa-}} & \multirow{2}{*}[9mm]{\rotatebox[origin=c]{90}{CLe~\cite{rajagopalan2025awracle}}} &
    & & &
    \includegraphics[width=\fgsize\textwidth]{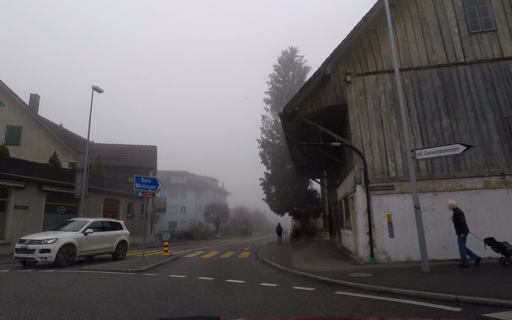} &
    \includegraphics[width=\fgsize\textwidth]{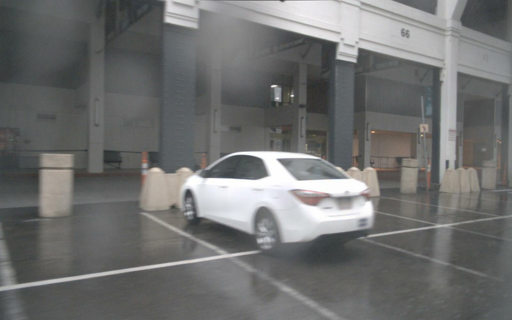} &
    \includegraphics[width=\fgsize\textwidth]{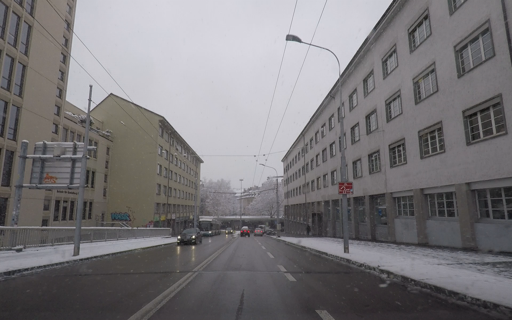}\\

    \multirow{2}{*}[9mm]{\rotatebox[origin=c]{90}{\small BAGEL}} & \multirow{2}{*}[7mm]{\rotatebox[origin=c]{90}{\cite{deng2025emerging}}} & 
    \includegraphics[width=\fgsize\textwidth]{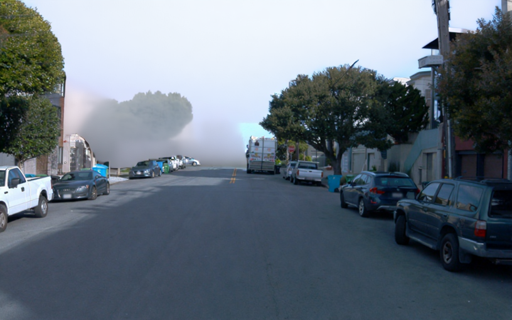} &
    \includegraphics[width=\fgsize\textwidth]{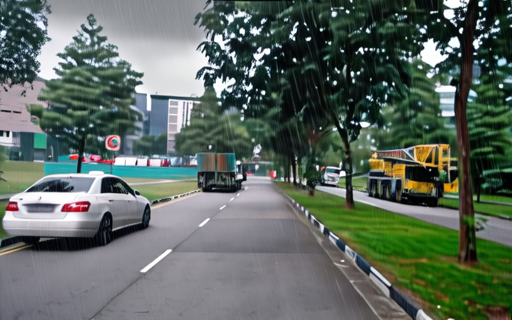} &
    \includegraphics[width=\fgsize\textwidth]{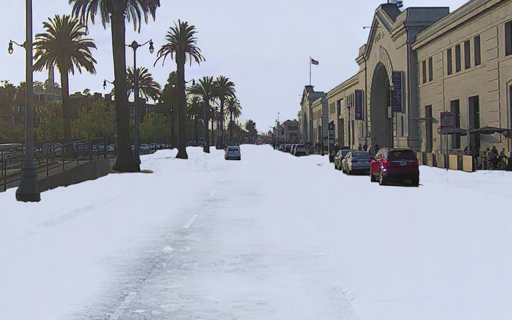} &
    \includegraphics[width=\fgsize\textwidth]{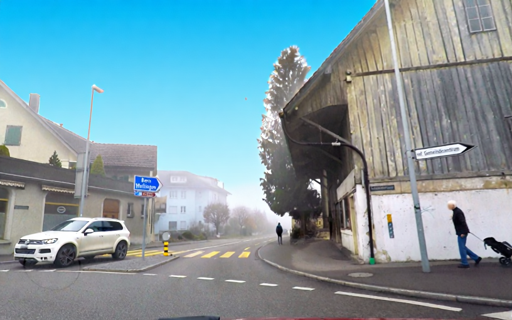} &
    \includegraphics[width=\fgsize\textwidth]{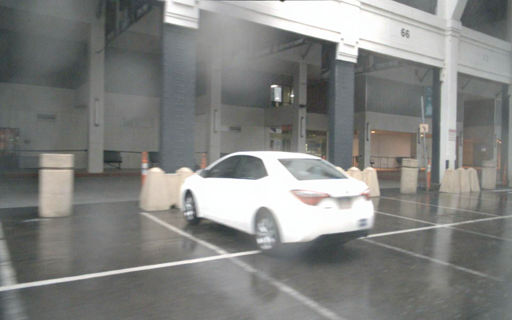} &
    \includegraphics[width=\fgsize\textwidth]{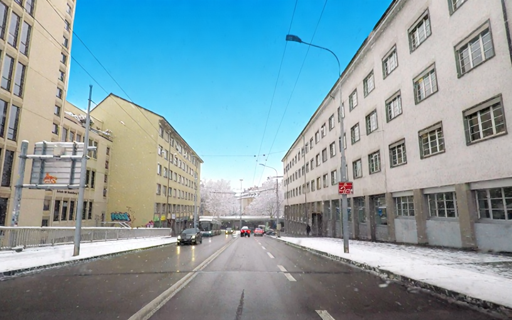}\\
    
    \multirow{2}{*}[9mm]{\rotatebox[origin=c]{90}{\small Qwen-I-E}} & \multirow{2}{*}[7mm]{\rotatebox[origin=c]{90}{\cite{bai2025qwen2}}} & 
    \includegraphics[width=\fgsize\textwidth]{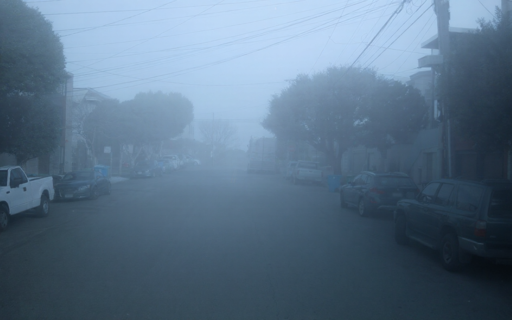} &
    \includegraphics[width=\fgsize\textwidth]{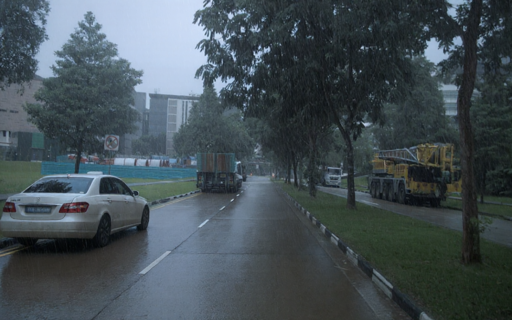} &
    \includegraphics[width=\fgsize\textwidth]{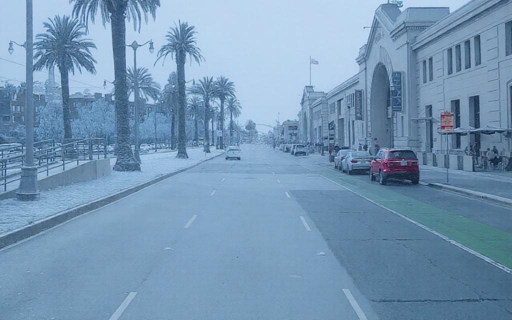} &
    \includegraphics[width=\fgsize\textwidth]{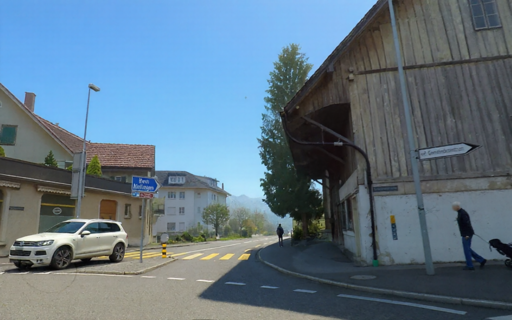} &
    \includegraphics[width=\fgsize\textwidth]{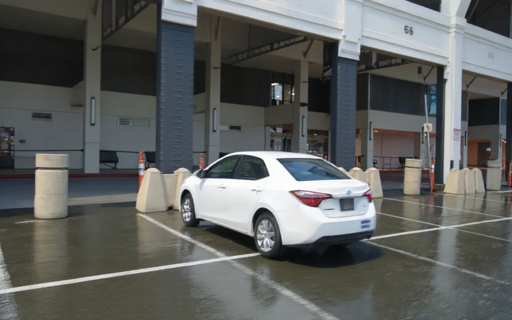} &
    \includegraphics[width=\fgsize\textwidth]{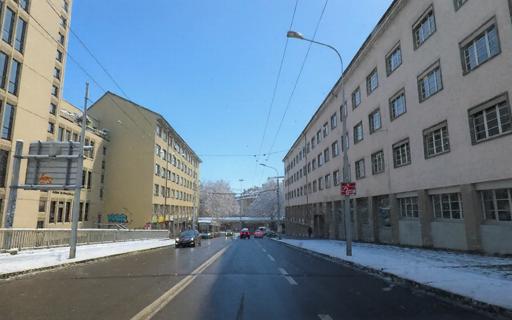}\\

    \multirow{2}{*}[8mm]{\rotatebox[origin=c]{90}{\small Token-}} & \multirow{2}{*}[9mm]{\rotatebox[origin=c]{90}{flow~\cite{geyer2023tokenflow}}} &
    \includegraphics[width=\fgsize\textwidth]{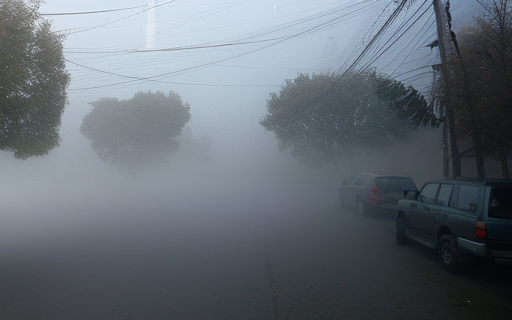} &
    \includegraphics[width=\fgsize\textwidth]{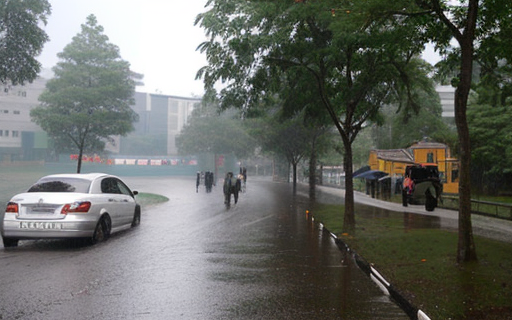} &
    \includegraphics[width=\fgsize\textwidth]{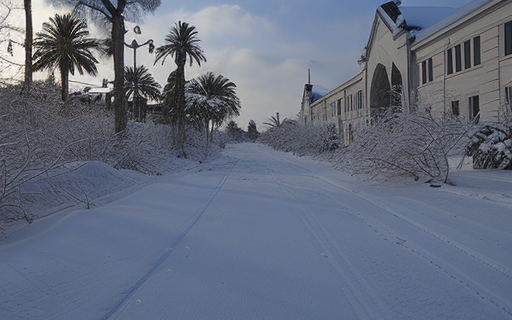} &
    \includegraphics[width=\fgsize\textwidth]{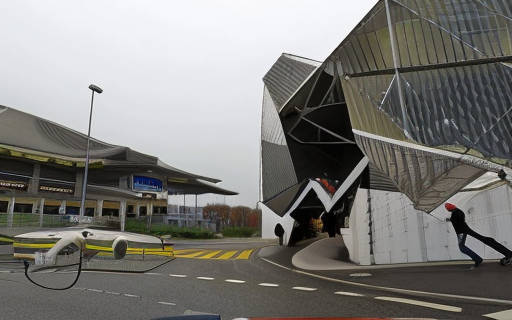} &
    \includegraphics[width=\fgsize\textwidth]{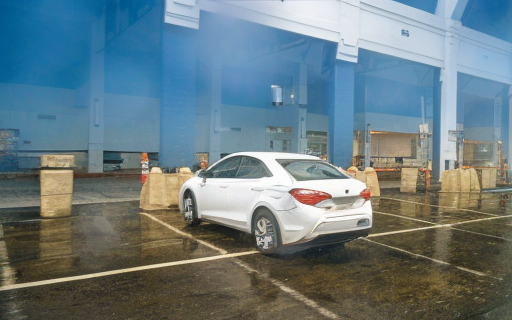} &
    \includegraphics[width=\fgsize\textwidth]{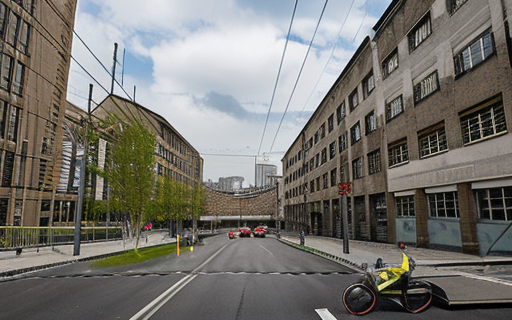}\\

    \multirow{2}{*}[9mm]{\rotatebox[origin=c]{90}{\small CycleGAN}} & \multirow{2}{*}[7mm]{\rotatebox[origin=c]{90}{\cite{zhu2017unpaired}}} &
    \includegraphics[width=\fgsize\textwidth]{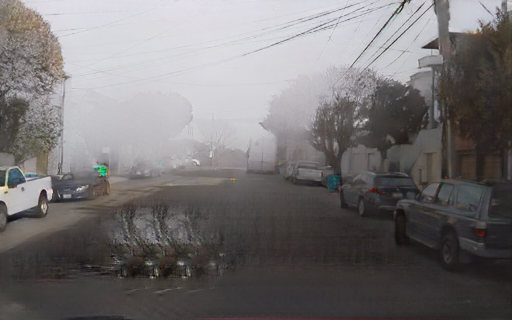} &
    \includegraphics[width=\fgsize\textwidth]{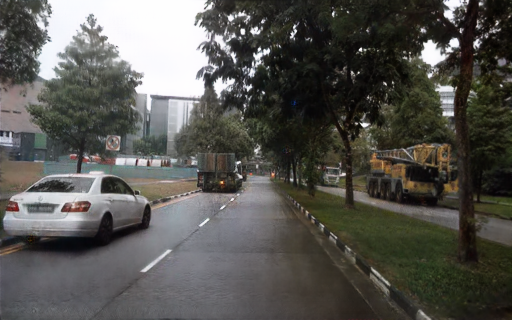} &
    \includegraphics[width=\fgsize\textwidth]{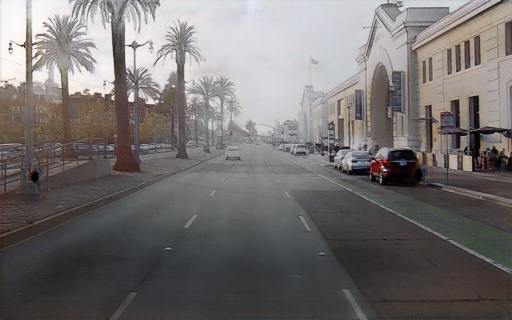} &
    \includegraphics[width=\fgsize\textwidth]{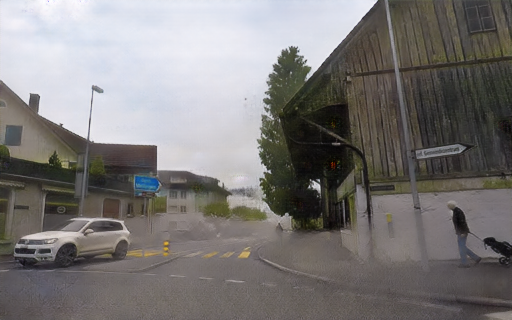} &
    \includegraphics[width=\fgsize\textwidth]{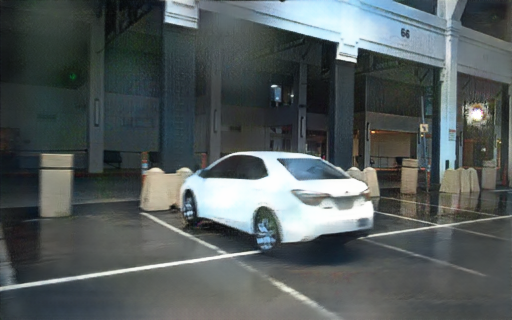} &
    \includegraphics[width=\fgsize\textwidth]{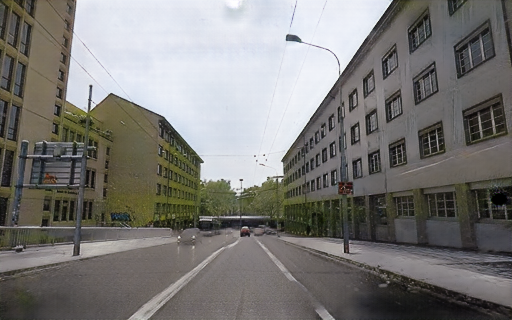}\\
    
    \multirow{2}{*}[7mm]{\rotatebox[origin=c]{90}{\small CycleNet}} & \multirow{2}{*}[7mm]{\rotatebox[origin=c]{90}{\cite{xu2023cyclenet}}} &
    \includegraphics[width=\fgsize\textwidth]{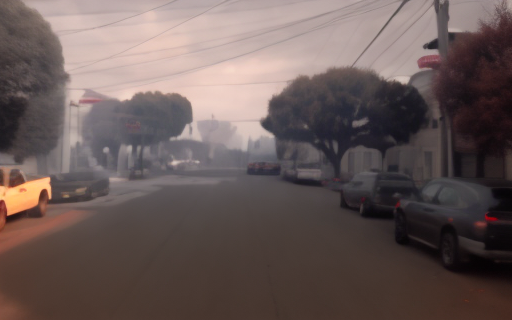} &
    \includegraphics[width=\fgsize\textwidth]{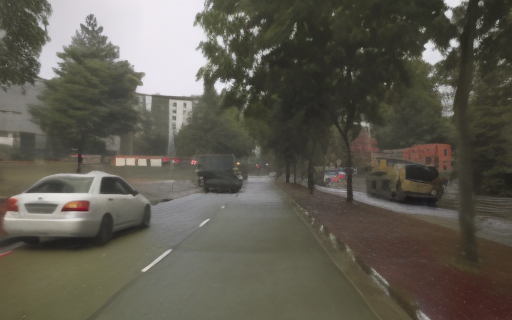} &
    \includegraphics[width=\fgsize\textwidth]{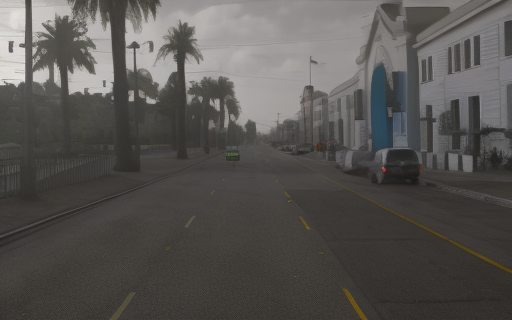} &
    \includegraphics[width=\fgsize\textwidth]{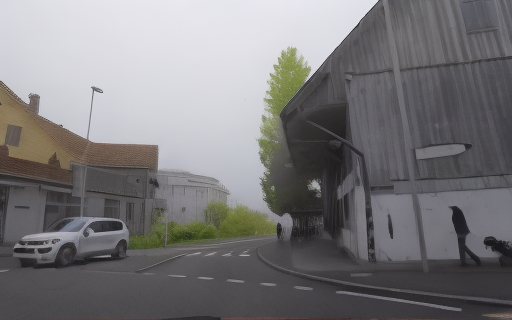} &
    \includegraphics[width=\fgsize\textwidth]{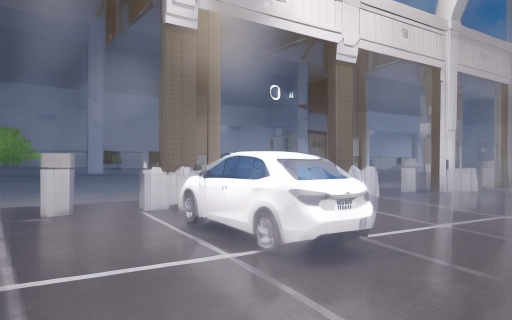} &
    \includegraphics[width=\fgsize\textwidth]{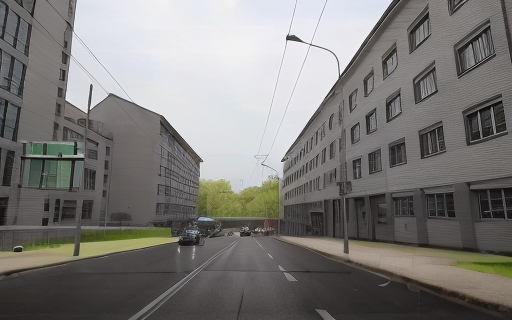}\\

    \multirow{2}{*}[7mm]{\rotatebox[origin=c]{90}{Cyclone}} & \multirow{2}{*}[8mm]{\rotatebox[origin=c]{90}{(ours)}} &
    \includegraphics[width=\fgsize\textwidth]{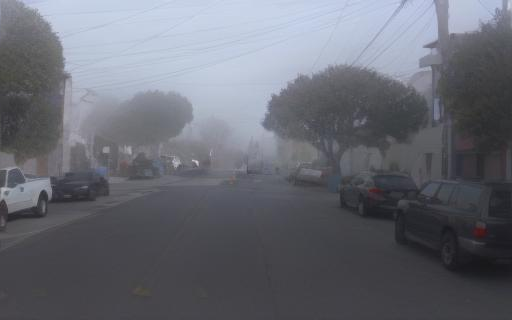} &
    \includegraphics[width=\fgsize\textwidth]{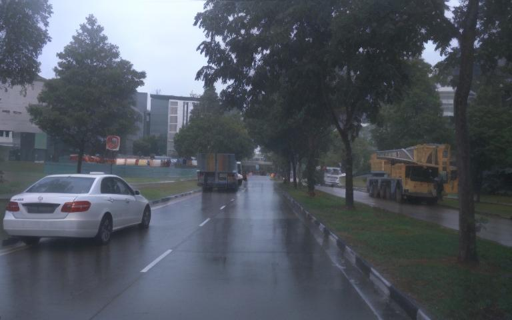} &
    \includegraphics[width=\fgsize\textwidth]{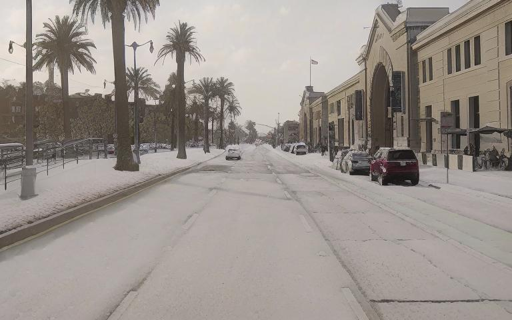} &
    \includegraphics[width=\fgsize\textwidth]{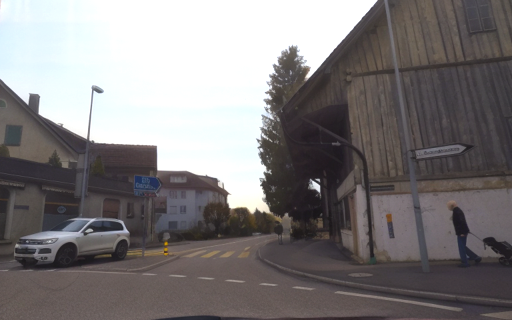} &
    \includegraphics[width=\fgsize\textwidth]{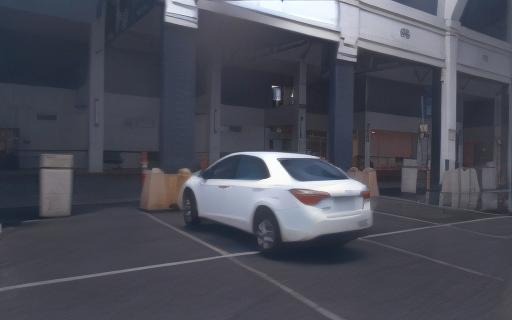} &
    \includegraphics[width=\fgsize\textwidth]{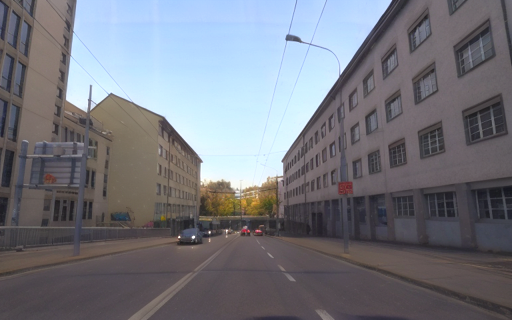}\\
    
    {} & {} & \multicolumn{3}{c}{Weather Synthesis} & \multicolumn{3}{c}{Weather Removal}\\
\end{tabular}
\vspace{-0.2cm}
\caption{Qualitative comparison with weather editing methods.}
\vspace{-1cm}
\label{fig:weather_synthesis}
\end{figure*} 

\begin{table*}
\caption{Quantitative comparison with various weather editing methods, highlighted with \hlc[red!30]{\textbf{best}}, \hlc[orange!30]{second}, and
\hlc[yellow!30]{third}.}
\centering
\resizebox{\textwidth}{!}{%
\begin{tabular}{l c cccc cccc}
    \toprule 
    \multirow{2}{*}{Method} & \multirow{2}{*}{\#Params} & \multicolumn{4}{c}{Weather Synthesis} & \multicolumn{4}{c}{Weather Removal} \\
    \cmidrule(rl){3-6} 
    \cmidrule(rl){7-10}
    {} & {} & DINO$\downarrow$ & CLIP$\uparrow$ & Acc$_{cls}\uparrow$ & FID$\downarrow$ & DINO$\downarrow$ & CLIP$\uparrow$ & Acc$_{cls}\uparrow$ & FID$\downarrow$\\
    \midrule
    Histoformer~\cite{sun2024restoring} &  0.2B & - & - & - & - & \sbest{0.469} & 20.45 & 0.00 & 101.08\\
    AWRaCLe~\cite{rajagopalan2025awracle} &  0.3B & - & - & - & - & \best{\textbf{0.344}} & 20.54 & 0.00 & 102.36\\
    \midrule
    InstructPix2Pix~\cite{brooks2023instructpix2pix} &  0.9B & \best{\textbf{2.065}} & \tbest{24.98} & \tbest{91.63} & 80.44 & \tbest{1.875} & \tbest{21.43} & 7.19 & 86.52\\
    BAGEL~\cite{deng2025emerging} & 7B & 3.195 & 23.60 & 74.20 & 70.13 & 1.887 & 21.24 & 8.61 & 86.59\\
    Qwen-Image-Edit~\cite{bai2025qwen2} & 20B & 2.848 & \best{\textbf{26.68}} & \best{\textbf{95.22}} & 68.04 & 2.483 & \sbest{21.72} & \sbest{43.80} & 83.24\\
    \midrule
    TokenFlow~\cite{geyer2023tokenflow} & 0.9B & 3.346 & 24.17 & 76.67 & 97.85 & 3.132 & 20.49 & 16.02 & 102.58\\
    CycleGAN~\cite{zhu2017unpaired} & 0.2B$\times$ 6 & \tbest{2.383} & 24.45 & 85.63 & \tbest{69.42} & 3.351 & 20.51 & \tbest{42.29} & \sbest{76.76}\\
    CycleNet~\cite{xu2023cyclenet} & 0.9B+0.4B$\times$ 3 & 2.396 & 24.04 & 78.81 & \sbest{68.05} & 2.184 & 20.59 & 13.36 & \tbest{79.65}\\
    Cyclone (ours) & 0.9B & \sbest{2.197} & \sbest{25.92} & \sbest{93.79} & \best{\textbf{62.58}} & 2.574 & \best{\textbf{22.30}} & \best{\textbf{92.80}} & \best{\textbf{65.96}}\\
    \bottomrule
\end{tabular}}
\label{tab:quantitative}
\end{table*}

\subsection{Qualitative Evaluation}
As illustrated in \cref{fig:weather_synthesis}, Histoformer and AWRaCLe, which are designed for removing synthetic rain and snow, generalize poorly to real-world weather and often produce outputs that are nearly identical to the input.
BAGEL and Qwen-Image-Edit can effectively add weather effects but struggle to remove existing ones, such as wet or snow-covered road surfaces. In particular, the results produced by BAGEL often appear less realistic.
TokenFlow, which relies on inversion-based editing, is prone to instability and may inadvertently alter the appearance of objects, making it difficult to preserve both scene fidelity and the desired weather condition.
CycleGAN and CycleNet rarely generate visually plausible results and frequently exhibit severe color shifts. In contrast, our method produces realistic and consistent edits across a wide range of weather conditions.

\subsection{Ablation Study}
\begin{figure}
\centering    
\begin{subfigure}{0.165\linewidth}
    \includegraphics[width=\linewidth]{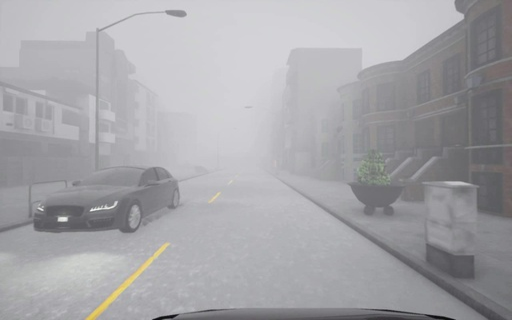}
    \caption*{Input}
\end{subfigure}
\begin{subfigure}{0.165\linewidth}
    \includegraphics[width=\linewidth]{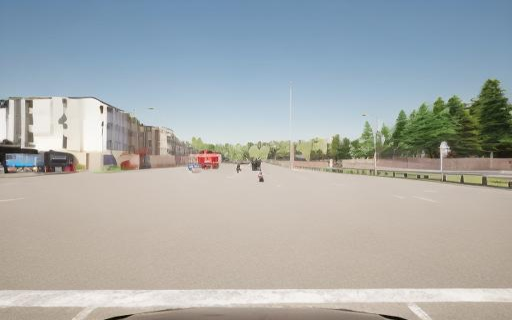}
    \caption*{w/o recon,cycle}
\end{subfigure}
\begin{subfigure}{0.165\linewidth}
    \includegraphics[width=\linewidth]{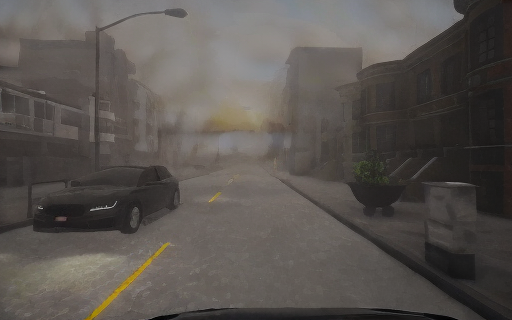}
    \caption*{w/o recon}
\end{subfigure}
\begin{subfigure}{0.165\linewidth}
    \includegraphics[width=\linewidth]{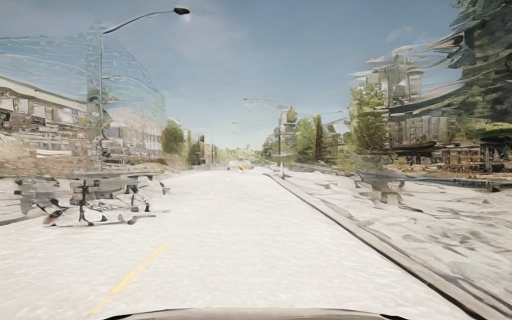}
    \caption*{w/o cycle}
\end{subfigure}\\
\begin{subfigure}{0.165\linewidth}
    \includegraphics[width=\linewidth]{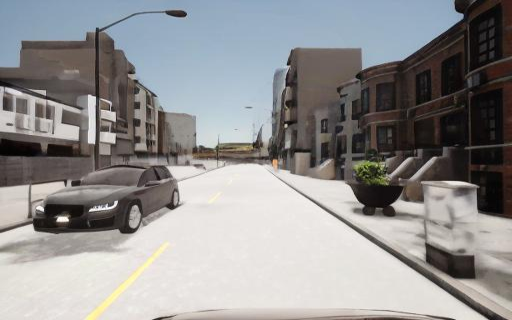}
    \caption*{Full model}
\end{subfigure}
\begin{subfigure}{0.165\linewidth}
    \includegraphics[width=\linewidth]{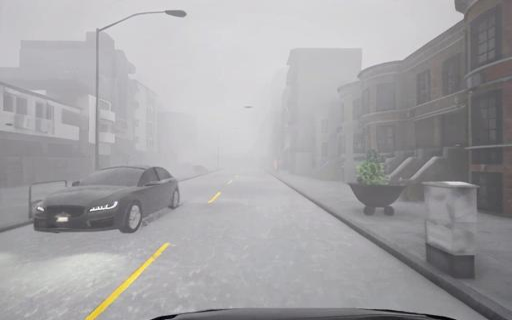}
    \caption*{w/o distill,CLIP}
\end{subfigure}
\begin{subfigure}{0.165\linewidth}
    \includegraphics[width=\linewidth]{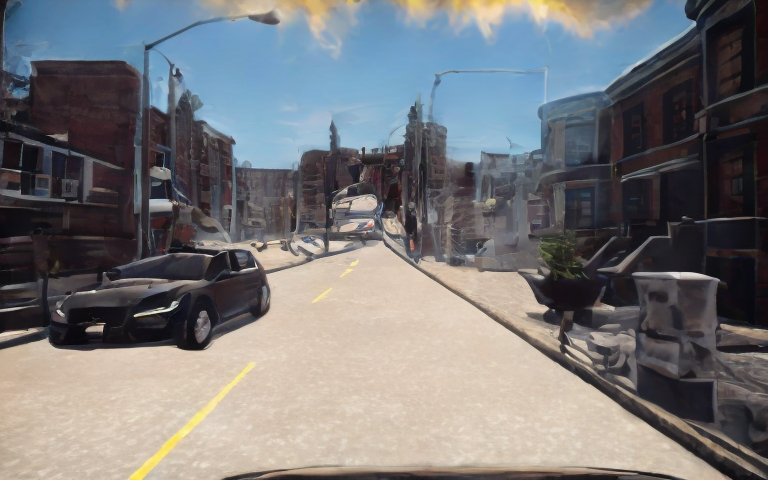}
    \caption*{w/o distill}
\end{subfigure}
\begin{subfigure}{0.165\linewidth}
    \includegraphics[width=\linewidth]{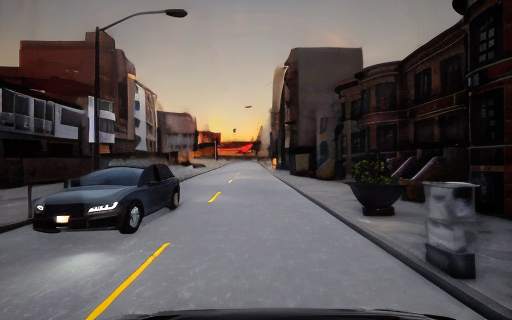}
    \caption*{w/o CLIP}
\end{subfigure}
\caption{\textbf{Ablation Study.} The full model effectively captures the scene and follows the target condition closely, while removing components leads to a loss in scene structure or reduced fidelity of the weather effects.}
\label{fig:ablation}
\vspace{-0.5cm}
\end{figure}

To study the design choices of \method, we experiment by removing each key component in turn. We train the model for 20k iterations on the SHIFT dataset to save computation. The results are presented in \cref{tab:ablation} and \cref{fig:ablation}.

\noindent\textbf{Cycle-Consistency.}
$\mathcal{L}_{recon}$ enhances details in the final output while $\mathcal{L}_{cycle}$ ensures the model preserves the original scene structure, preventing scene distortions.
Removing these constraints causes the model to lose its ability to preserve structural information from the input image. Consequently, the model relies solely on prior knowledge and textual conditions for generation, resulting in stable CLIP similarity and accuracy but significantly degraded DINO-Struct.

\noindent\textbf{Distillation and Textual Guidance.} 
The distillation step acts as an effective regularization mechanism, preventing overfitting. Without it, the model tends to jointly learn both the structure and the weather effects from the input, yielding good DINO-Struct but deteriorating other metrics. $\mathcal{L}_{distil}$ forces the model to follow the distribution of the training data.
Incorporating text prompts can further improve the effect, boosting the CLIP and accuracy metrics. Because CLIP represents an image using a single global embedding, it cannot capture fine-grained details as effectively as the teacher model. Therefore, $\mathcal{L}_{CLIP}$ alone steers the output toward the target prompt, but introduces unwanted artifacts. 

\noindent\textbf{The full model} combines all components, striking the best balance between structural consistency and textual fidelity and achieving strong performance across all metrics.

\subsection{Applications}
Our model is suitable for a range of applications, including data augmentation, simulation, and improving perception models.

\def\fgsize{0.18}

\begin{figure*}
\centering
\setlength{\tabcolsep}{0.002\linewidth}
\renewcommand{\arraystretch}{0.8}
\begin{tabular}{cccccc}
    \includegraphics[width=\fgsize\textwidth]{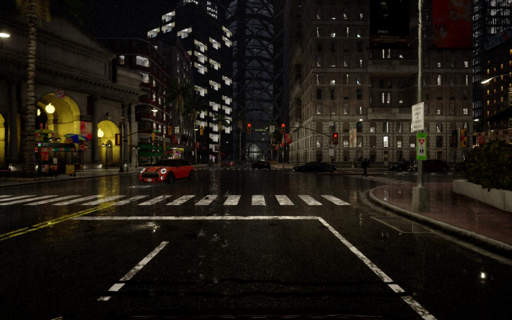} &
    \raisebox{6mm}{$\longrightarrow$} &
    \includegraphics[width=\fgsize\textwidth]{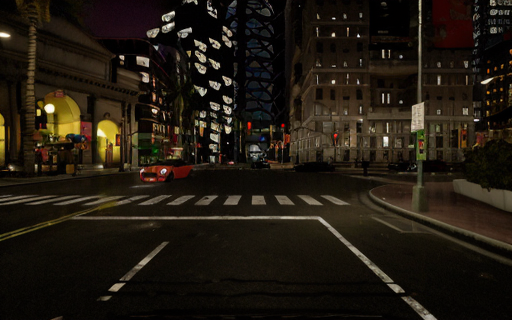} &
    \includegraphics[width=\fgsize\textwidth]{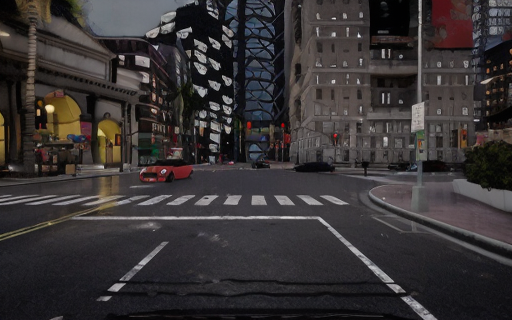} &
    \includegraphics[width=\fgsize\textwidth]{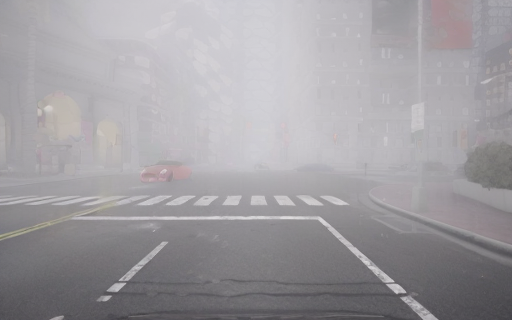} &
    \includegraphics[width=\fgsize\textwidth]{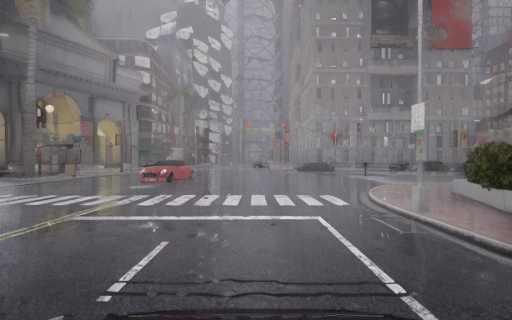}\\
    Rainy night (Input) & {} & Clear night & Clear day & Foggy day & Rainy day\\
    
    \includegraphics[width=\fgsize\textwidth]{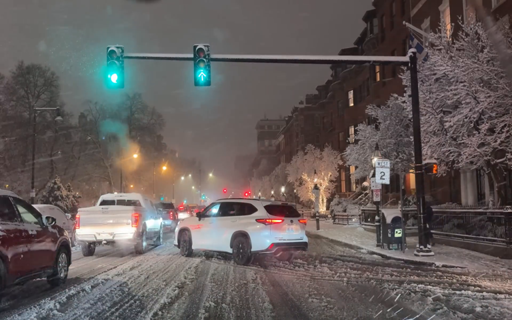} &
    \raisebox{6mm}{$\longrightarrow$} &
    \includegraphics[width=\fgsize\textwidth]{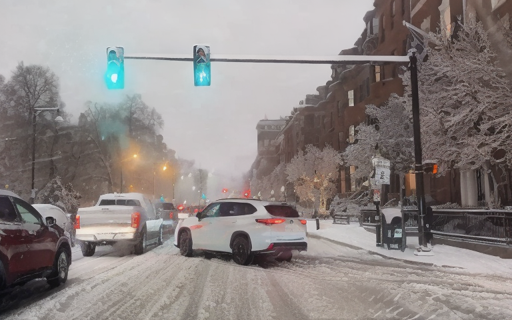} &
    \includegraphics[width=\fgsize\textwidth]{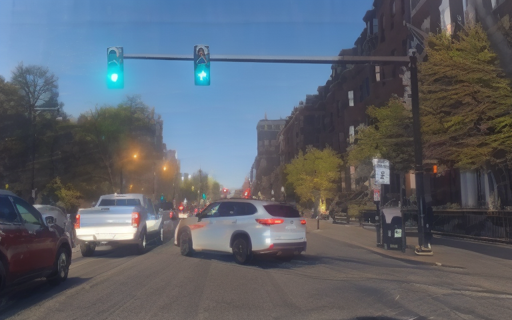} &
    \includegraphics[width=\fgsize\textwidth]{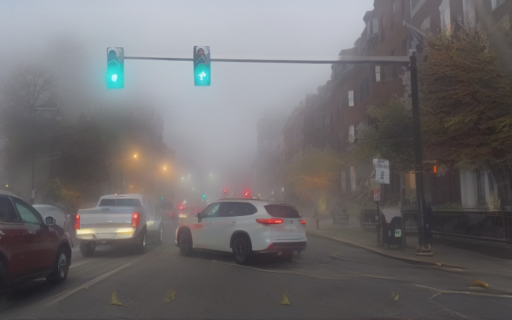} &
    \includegraphics[width=\fgsize\textwidth]{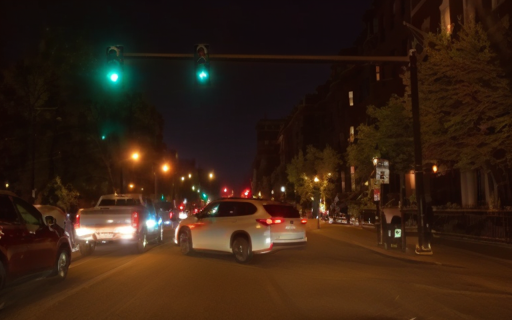}\\
    Snowy night (Input) & {} & Snowy day & Clear day & Foggy day & Clear night\\
    
    \includegraphics[width=\fgsize\textwidth]{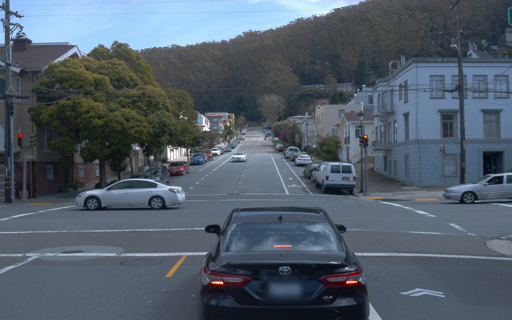} &
    \raisebox{6mm}{$\longrightarrow$} &
    \includegraphics[width=\fgsize\textwidth]{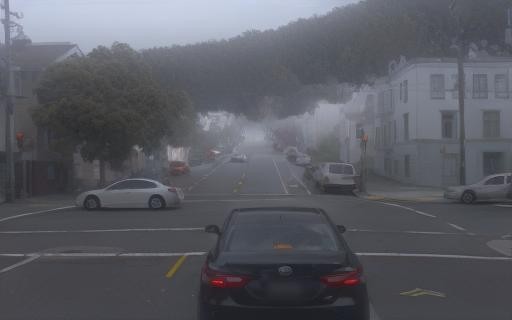} &
    \includegraphics[width=\fgsize\textwidth]{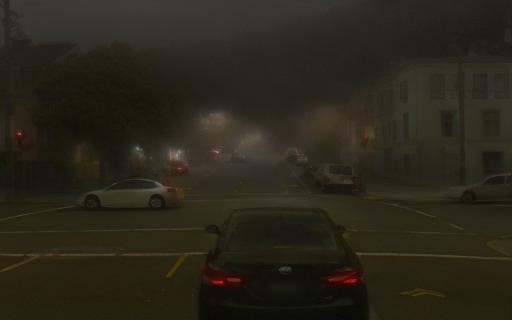} &
    \includegraphics[width=\fgsize\textwidth]{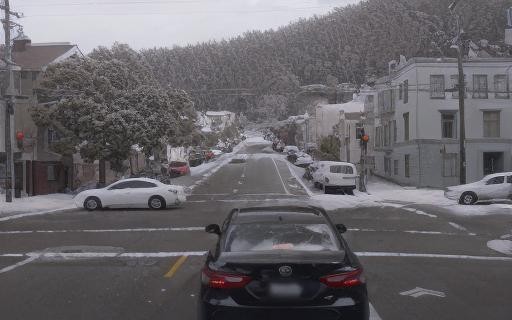} &
    \includegraphics[width=\fgsize\textwidth]{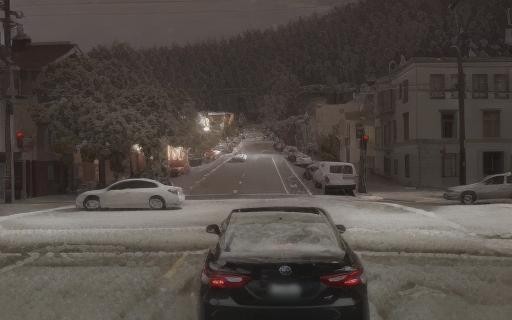} \\
    Clear day (Input)& {} & Foggy day & Foggy night & Snowy day & Snowy night\\
\end{tabular}
\caption{\textbf{Multiple Weathers Editing.} Without focusing solely on adding or removing weather, Cyclone enables editing across diverse weathers and between times of day.}
\label{fig:multi_weather}
\vspace{-0.3cm}
\end{figure*}

\noindent\textbf{Video Diffusion Fine-tuning.}
Driving logs are typically captured as sequential data over multiple time steps. Ideally, weather effects and scene elements should remain consistent across temporal and spatial dimensions. However, treating each image independently can introduce inconsistencies across frames and viewpoints due to the inherent stochasticity of generative models. 
To address this limitation, we generate pseudo-paired data using the image model trained in Sec 4. to fine-tune a video diffusion model. Specifically, for each video input  $\mathbf{X}_0$ we generate a corresponding output $\hat{\mathbf{Y}}_0$. As the image model generates frames independently, inconsistencies across time steps can arise. We therefore use the generated frames as conditional inputs while supervising the video model with the ground-truth video.
The video model is optimized using the objective: $\mathcal{L}=\mathbb{E}_{t}\|D_{\theta}(\mathbf{X}_t,c_x, \hat{\mathbf{Y}}_0)-\mathbf{X}_0\|^2_2.$
This approach enables the model to leverage high-quality supervision while remaining robust to imperfect conditional inputs, thereby improving temporal consistency without compromising visual fidelity (see \cref{tab:temporal_consisency}). Additional video results are provided in the Supplementary Material.

\begin{table*}
\caption{Quantitative comparison of temporally consistent video editing.
We adopt VBench~\cite{huang2024vbench}, which evaluates Temporal Flickering (TF), Motion Smoothness (MS), Overall Consistency (OC), and FVD~\cite{unterthiner2019fvd}. We also report gains compared to Cyclone, indicating the temporal benefit when finetuning on video data.}
\centering
\resizebox{0.8\textwidth}{!}{%
\begin{tabular}{l cccc cccc}
    \toprule 
    \multirow{2}{*}{Method} & \multicolumn{4}{c}{Weather Synthesis} & \multicolumn{4}{c}{Weather Removal} \\
    \cmidrule(rl){2-5} 
    \cmidrule(rl){6-9}
    {} & FVD$\downarrow$ & TF$\uparrow$ & MS$\uparrow$ & OC$\uparrow$ & FVD$\downarrow$ & TF$\uparrow$ & MS$\uparrow$ & OC$\uparrow$\\
    \midrule
    Cyclone & 1225 & 94.32 & 95.65 & \best{\textbf{20.42}} & \sbest{1557} & 93.94 & 95.18 & \best{\textbf{15.91}}\\
    \midrule
    TokenFlow~\cite{geyer2023tokenflow} & \sbest{1162} & \sbest{96.42} & \sbest{97.64} & 19.20 & 2040 & \sbest{96.82} & \sbest{97.72} & 12.08\\
    SVD-ft~\cite{blattmann2023stable} w/ Cyclone & \best{\textbf{1000}} & \best{\textbf{96.88}} & \best{\textbf{98.53}} & \sbest{19.52} & \best{\textbf{1161}} & \best{\textbf{97.04}} & \best{\textbf{98.06}} & \sbest{14.90}\\
    \midrule
    Gain & 225 & 2.56 & 2.88 & -0.90 & 396 & 3.10 & 2.88 & -1.01\\
    \bottomrule
\end{tabular}}
\label{tab:temporal_consisency}
\end{table*}

\noindent\textbf{One-to-Many Weather Synthesis.}
We do not restrict our method to merely adding or removing effects; instead, we enable adaptive transitions among different weather conditions.
Furthermore, we extend our text conditioning to incorporate “daytime” and “nighttime”, expanding the weather set to eight distinct conditions. This extension allows our model to synthesize weather effects under nighttime settings as shown in \cref{fig:multi_weather}.

\begin{table}
\begin{minipage}[t]{.44\textwidth}
    \caption{\textbf{Quantitative ablation study.} We ablate each loss component during training.}
    \centering
    \resizebox{\columnwidth}{!}{%
    \begin{tabular}{l|cccc}
        \toprule
         Ablation & DINO$\downarrow$ & CLIP$\uparrow$ & Acc$_{cls}\uparrow$ & FID$\downarrow$\\
        \midrule
        w/o $\mathcal{L}_{recon}, \mathcal{L}_{cycle}$ & 9.476 & 17.36 & \best{\textbf{93.10}} & \best{\textbf{60.93}}\\
        w/o $\mathcal{L}_{recon}$ & 4.496 & \tbest{18.57} & 78.04 & 106.67\\
        w/o $\mathcal{L}_{cycle}$ & 5.569 & 18.36 & 77.76 & 121.46\\
        w/o $\mathcal{L}_{distill},\mathcal{L}_{clip}$ & \best{\textbf{0.577}} & 16.95 & 2.52 & 110.30\\
        w/o $\mathcal{L}_{distill}$ & 4.566 & \sbest{18.93} & \sbest{87.18} & 108.67\\
        w/o $\mathcal{L}_{clip}$ & \sbest{3.141} & 18.29 & 80.35 & \tbest{73.11}\\
        \midrule
        Full model & \tbest{3.244} & \best{\textbf{18.95}} & \tbest{81.72} & \sbest{70.92}\\
        \bottomrule
    \end{tabular}}
    \label{tab:ablation}
\end{minipage}
\hfill
\begin{minipage}[t]{.54\textwidth}
    \caption{Quantitative evaluation of multiple perception tasks on degraded scenes with and without removing adverse effects.}
    \resizebox{\columnwidth}{!}{%
    \begin{tabular}{l|ccccc}
        \toprule
         SHIFT~\cite{sun2022shift} & $\delta_1\uparrow$ & AbsRel$\downarrow$ & RMSE$\downarrow$ & mIoU$\uparrow$ & mAP$\uparrow$\\
        \midrule
        w/o weather removal & 51.70 & 0.298 & 7.288 & 35.61 & 24.93\\
        w/ weather removal & \textbf{58.61} & \textbf{0.260} & \textbf{6.613} & \textbf{37.47} & \textbf{28.86}\\
        \bottomrule
    \end{tabular}}
    \resizebox{0.72\columnwidth}{!}{%
    \begin{tabular}{l|ccccc}
        \toprule
        ACDC~\cite{sakaridis2021acdc} & mIoU$\uparrow$ & mAP$\uparrow$ & AP$_{50}\uparrow$\\
        \midrule
        w/o weather removal & 44.67 & 26.36 & 34.86\\
        w/ weather removal & \textbf{47.83} &  \textbf{32.35} & \textbf{41.07}\\
        \bottomrule
    \end{tabular}}
    \label{tab:downstream}
\end{minipage}
\end{table}

\begin{figure}
\centering    
\begin{subfigure}{\linewidth}
\centering
    \includegraphics[width=0.165\linewidth]{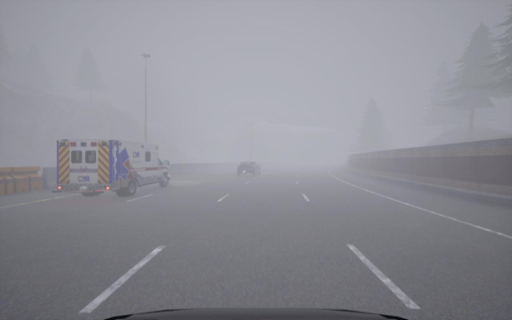}
    \includegraphics[width=0.165\linewidth]{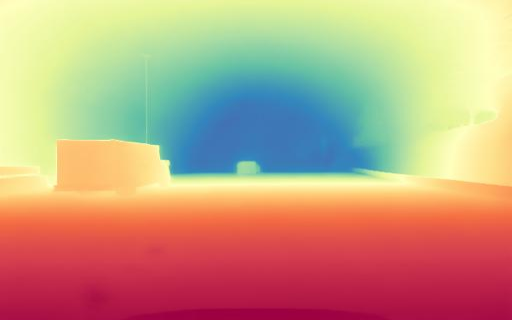}
    \includegraphics[width=0.165\linewidth]{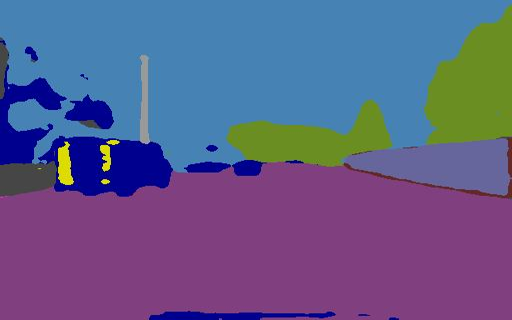}
    \includegraphics[width=0.165\linewidth]{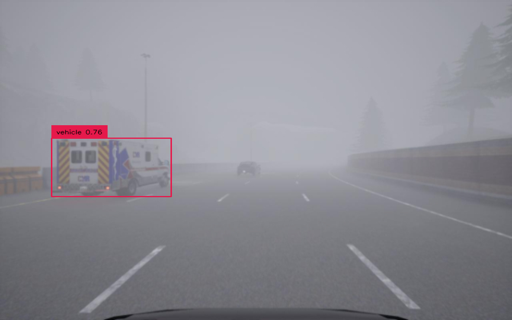}\\
    \includegraphics[width=0.165\linewidth]{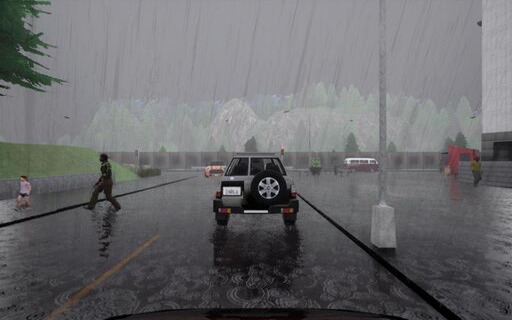}
    \includegraphics[width=0.165\linewidth]{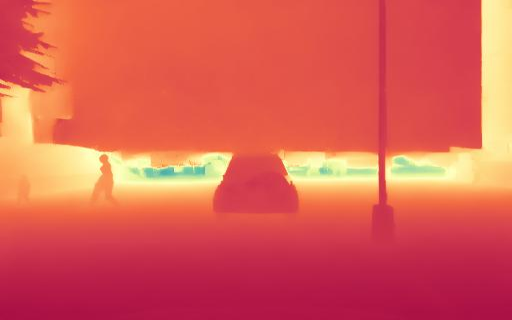}
    \includegraphics[width=0.165\linewidth]{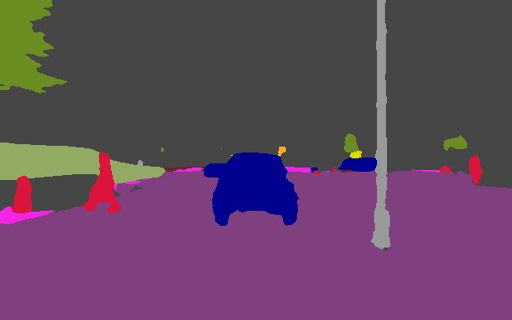}
    \includegraphics[width=0.165\linewidth]{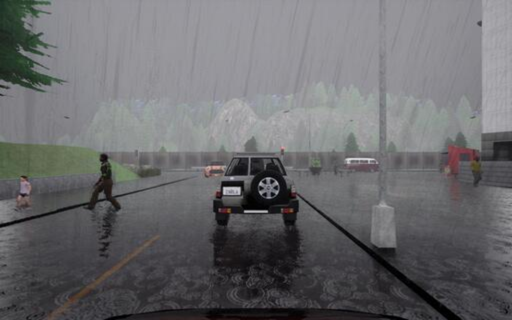}
    \caption*{w/o weather removal}
\end{subfigure}

\begin{subfigure}{\linewidth}
\centering
    \includegraphics[width=0.165\linewidth]{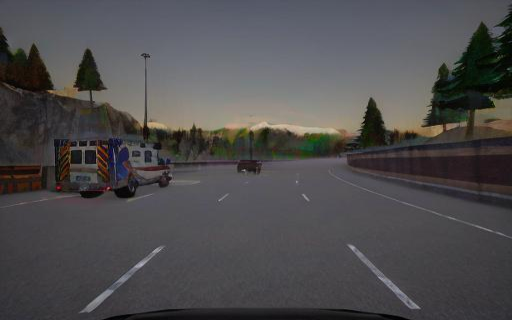}
    \includegraphics[width=0.165\linewidth]{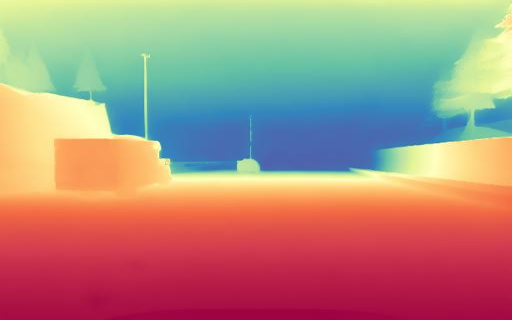}
    \includegraphics[width=0.165\linewidth]{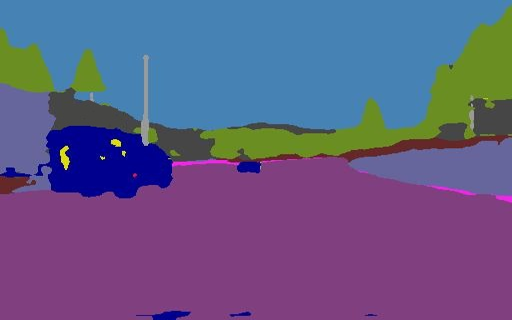}
    \includegraphics[width=0.165\linewidth]{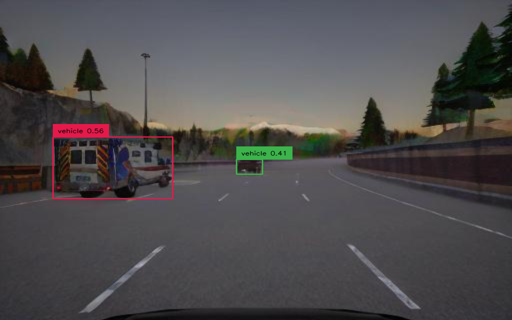}\\
    \includegraphics[width=0.165\linewidth]{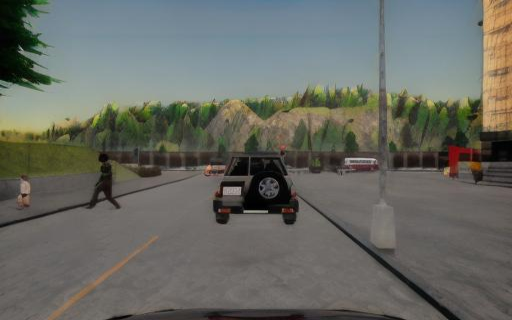}
    \includegraphics[width=0.165\linewidth]{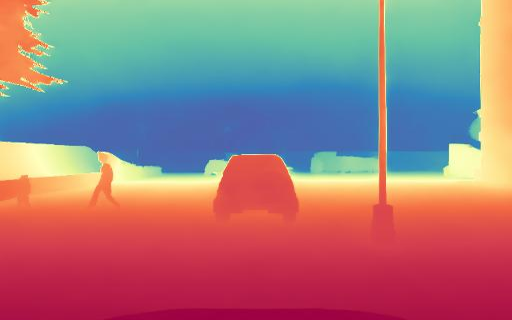}
    \includegraphics[width=0.165\linewidth]{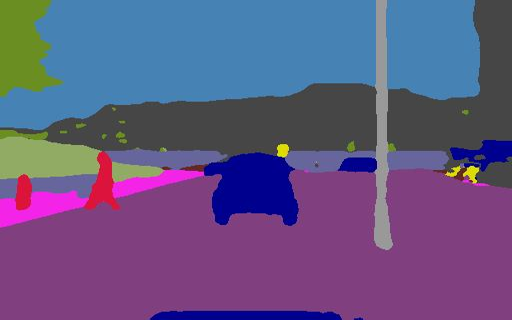}
    \includegraphics[width=0.165\linewidth]{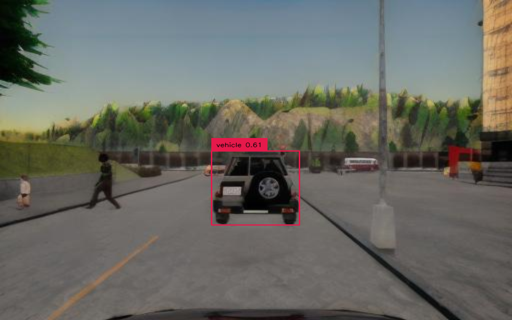}
    \caption*{w/ weather removal}
\end{subfigure}
\caption{\textbf{Improved perception tasks with weather removal.} 
Left to right: images used, estimated depth maps, segmentation maps, detected objects.}
\label{fig:downstream}
\vspace{-1cm}
\end{figure}

\noindent\textbf{Downstream Tasks.}
We evaluate the effectiveness of our method on removing bad weather and its impact on downstream perception tasks. Specifically, we conduct experiments on SHIFT~\cite{sun2022shift} and ACDC~\cite{sakaridis2021acdc}, which provide ground-truth annotations under a wide range of adverse conditions. We consider zero-shot depth estimation using Marigold~\cite{ke2024repurposing}, semantic segmentation using Mask2Former~\cite{cheng2022masked}, and object detection using Grounding  DINO~\cite{liu2024grounding}. We fix the target prompt to “clear daytime” and report results before and after applying our method.
As shown in \cref{tab:downstream} and \cref{fig:downstream}, removing adverse weather with our method consistently improves performance across all tasks, demonstrating its utility for enhancing perception under challenging conditions.

\subsection{Limitations}
Several limitations remain in our method. First, cycle training requires multiple forward passes per iteration, which increases training complexity and computations. Second, VAE compression and diffusion artifacts can lead to the loss of fine details (e.g., text and thin structures). Finally, our model cannot currently combine multiple weather effects with varying intensities, primarily due to the limited availability of annotated data.

\section{Conclusion}
In this work, we presented Cyclone, a unified framework for controllable weather synthesis. By leveraging latent diffusion models alongside cycle-consistency and knowledge distillation, our approach enables all-in-one weather editing while preserving scene structure and realism without any ground-truth pair. Experimental results demonstrate Cyclone outperforms existing baselines in both visual fidelity and structural consistency. We believe this framework provides a scalable and flexible solution for robust weather simulation, with potential applications in autonomous driving, robotics, and safety-critical perception research.

\bibliographystyle{splncs04}
\bibliography{main}

\clearpage
\appendix

\section{Implementation Details}
\noindent\textbf{Derivation of Invariance Bound.}
We provide the complete derivation of the timestep-dependent bound presented in Eq. 4 of the main paper. This bound demonstrates that invariance emerges implicitly from multi-domain cycle consistency without requiring an explicit invariance loss term.
Let $\mathcal{X}$ denote the source domain, and let $\mathcal{Y}$ and $\mathcal{Z}$ denote two arbitrary target weather domains. Given a source image $x_0 \in \mathcal{X}$, our model $D_\theta$ performs forward translations:
\begin{equation}
\bar{y} = D_\theta(x_t, c_y, x_0), \quad \bar{z} = D_\theta(x_t, c_z, x_0)
\end{equation}
We aim to bound the difference between outputs translated from two different domains back to $\mathcal{X}$. Consider the expected difference at timestep $t$:
\begin{equation}
\mathbb{E}_t \left\| D_\theta(x_t, c_x, \bar{y}) - D_\theta(x_t, c_x, \bar{z}) \right\|
\end{equation}
By adding and subtracting the ground truth source $x_0$, we apply the triangle inequality:
\begin{align}
\left\| D_\theta(x_t, c_x, \bar{y}) - D_\theta(x_t, c_x, \bar{z}) \right\| \nonumber &= \left\| \big(D_\theta(x_t, c_x, \bar{y}) - x_0\big) - \big(D_\theta(x_t, c_x, \bar{z}) - x_0\big) \right\| \nonumber \\
&\leq \left\| D_\theta(x_t, c_x, \bar{y}) - x_0 \right\| + \left\| D_\theta(x_t, c_x, \bar{z}) - x_0 \right\|
\label{eq:triangle}
\end{align}
Taking the expectation over timesteps $t\sim\mathcal{U}(0, T)$ on both sides:
\begin{equation}
\mathbb{E}_t \left\| D_\theta(x_t, c_x, \bar{y}) - D_\theta(x_t, c_x, \bar{z}) \right\| 
\leq \mathbb{E}_t \left\| D_\theta(x_t, c_x, \bar{y}) - x_0 \right\| 
+ \mathbb{E}_t \left\| D_\theta(x_t, c_x, \bar{z}) - x_0 \right\|
\end{equation}
Recall that our reconstruction losses are defined as squared $\ell_2$ norms:
\begin{align}
\mathcal{L}_{\text{recon}} &= \mathbb{E}_t \left\| D_\theta(x_t, c_x, \bar{y}) - x_0 \right\|_2^2 \\
\mathcal{L}'_{\text{recon}} &= \mathbb{E}_t \left\| D_\theta(x_t, c_x, \bar{z}) - x_0 \right\|_2^2
\end{align}
By Jensen's Inequality (since $f(x) = \sqrt{x}$ is concave, or equivalently 
$(\mathbb{E}[|X|])^2 \leq \mathbb{E}[|X|^2]$):
\begin{equation}
\mathbb{E}_t \left\| D_\theta(x_t, c_x, \bar{y}) - x_0 \right\| 
\leq \sqrt{\mathbb{E}_t \left\| D_\theta(x_t, c_x,, \bar{y}) - x_0 \right\|_2^2} 
= \sqrt{\mathcal{L}_{\text{recon}}}
\end{equation}
Combining the above results yields:
\begin{equation}
\boxed{
\mathbb{E}_t \left\| D_\theta(x_t, c_x, \bar{y}) - D_\theta(x_t, c_x, \bar{z}) \right\| 
\leq \sqrt{\mathcal{L}_{\text{recon}}} + \sqrt{\mathcal{L}'_{\text{recon}}} 
\triangleq B_t
}
\label{eq:bound_final}
\end{equation}

\subsection{Data Processing}
\begin{figure}
\centering    
\begin{subfigure}{0.49\linewidth}
    \includegraphics[width=0.32\linewidth]{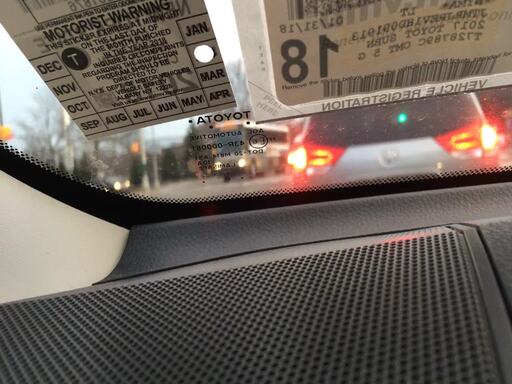}
    \includegraphics[width=0.32\linewidth]{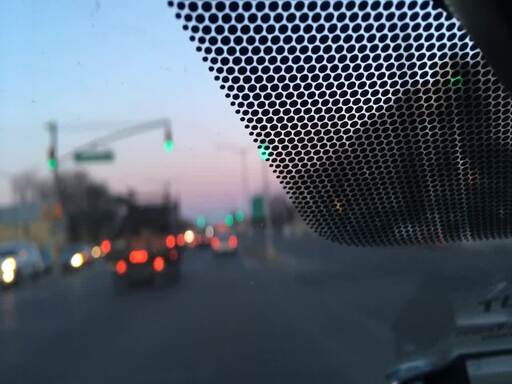}
    \includegraphics[width=0.32\linewidth]{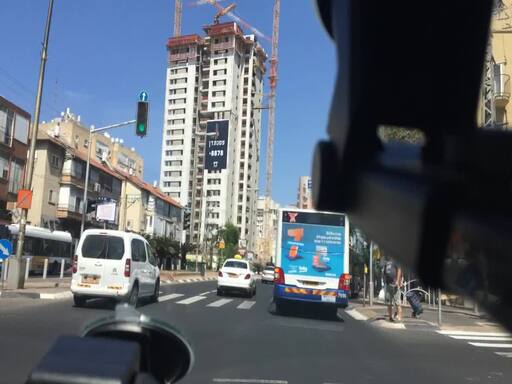}
    \caption*{High occlusion}
\end{subfigure}
\begin{subfigure}{0.16\linewidth}
    \includegraphics[width=\linewidth]{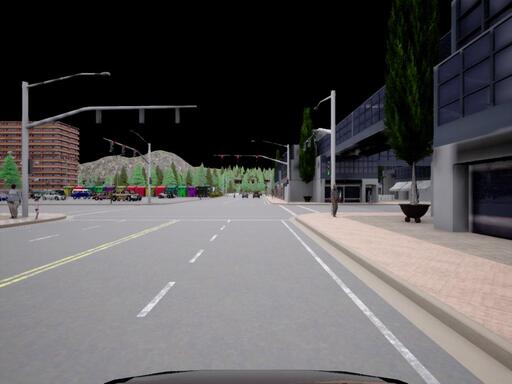}
    \caption*{Badly rendered}
\end{subfigure}
\begin{subfigure}{0.33\linewidth}
    \includegraphics[width=0.49\linewidth]{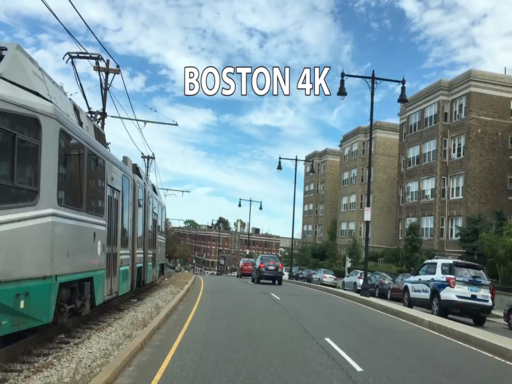}
    \includegraphics[width=0.49\linewidth]{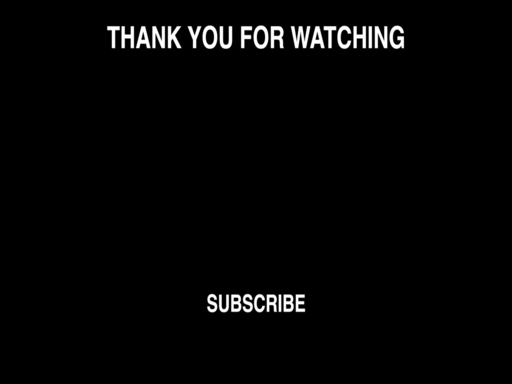}
    \caption*{Including texts}
\end{subfigure}
\caption{Samples that were not included during training.}
\label{fig:discard_data}
\end{figure}

For ACDC, SHIFT, and BDD100K, we use the official weather annotations provided with each dataset. We manually assign a weather label to each video in OpenDV-YouTube. We further filter out ambiguous or low-quality samples. Examples of the excluded scenes are shown in \cref{fig:discard_data}.
Because the resulting dataset is highly imbalanced across weather categories (sunny scenes are far more common than fog or snow), we adopt a weighted sampling strategy during training. This ensures that rare weather conditions are adequately represented.

\subsection{Training}
\begin{figure}
\centering    
\includegraphics[width=0.2\linewidth]{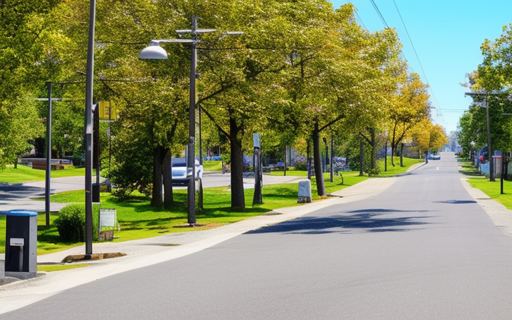}
\includegraphics[width=0.2\linewidth]{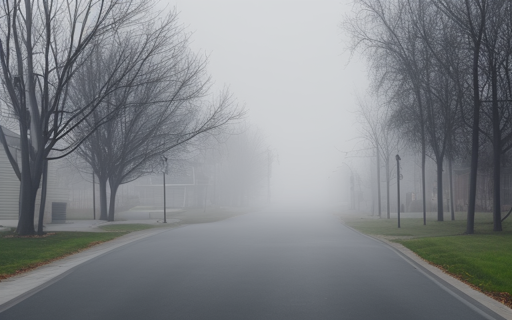}
\includegraphics[width=0.2\linewidth]{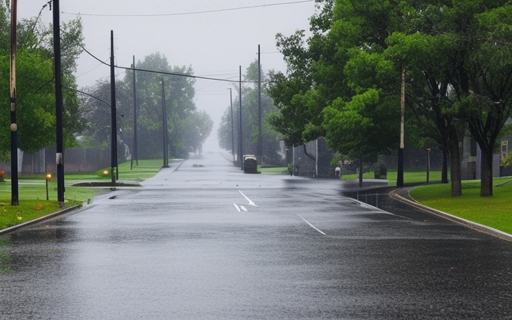}
\includegraphics[width=0.2\linewidth]{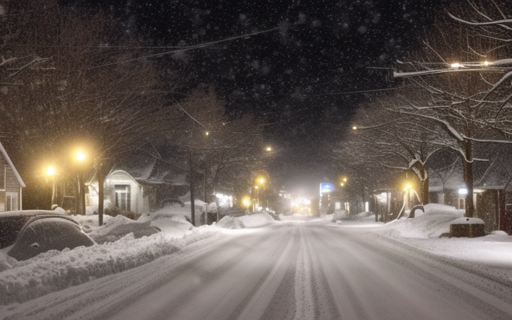}\\
\includegraphics[width=0.2\linewidth]{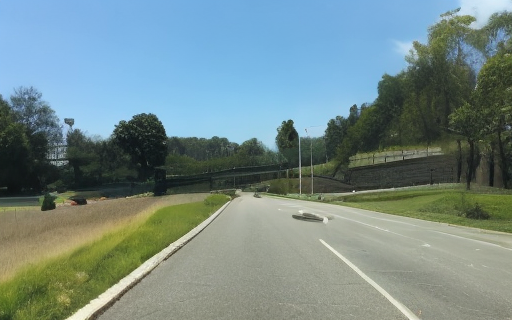}
\includegraphics[width=0.2\linewidth]{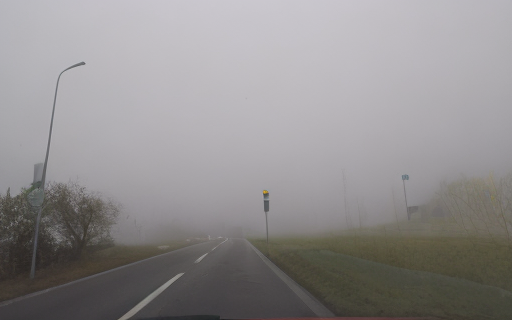}
\includegraphics[width=0.2\linewidth]{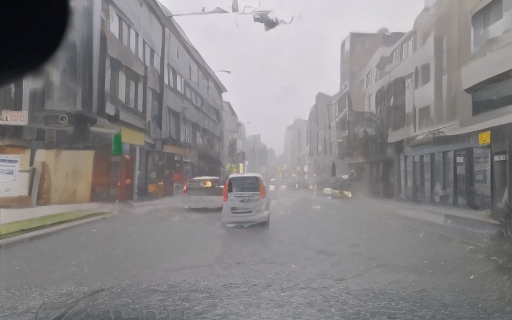}
\includegraphics[width=0.2\linewidth]{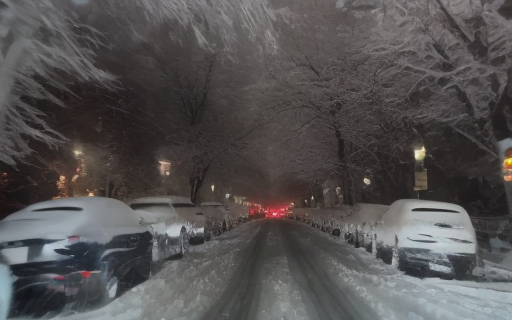}
\caption{Comparing samples generated with Stable Diffusion v2.1 (top), our adapted model (bottom). Fine-tuning increases the weather-specific realism by a margin. For the original SD v2.1 sampling, we used the text "a photo of a suburban street under $[...]$ weather".}
\label{fig:preadapted_compare}
\end{figure}

\noindent\textbf{Pre-adapted LDM.}
While our Stable Diffusion backbone is pre-trained on large-scale datasets such as LAION, which provide diverse and high-quality visual content, it is not directly suited for modeling the physical properties of driving-world weather conditions for distillation. To adapt the model to our task, we first fine-tune it using conventional text-to-image training driven by weather-specific prompts. The resulting fine-tuned model then serves as the initialization for both the student and teacher networks in the subsequent training stage.
To verify that fine-tuning effectively adapts the model to the target visual domain, we compare samples generated by the adapted model with those produced by the original SD v2.1 model. We report the results in \cref{fig:preadapted_compare}, which shows that the adapted model generates images that are visually more consistent with the target domain and more closely resemble the training images.

\noindent\textbf{Training Details.}
The model is trained with a batch size of 8 and a learning rate of 1e-5 for 100k iterations. Other training parameters are set to their default values.
It takes 2 GPU days for training on GPUs equivalent to NVIDIA RTX 4090. During inference, we set CFG to 1 and perform denoising with 20 steps DDIM~\cite{song2020denoising}.

\subsection{Convergence Analysis}
\begin{figure*}
\centering
\begin{tabular}{ccccccc}
    & \multirow{2}{*}[10mm]{\rotatebox{90}{CycleNet}} 
    & \includegraphics[width=0.15\linewidth]{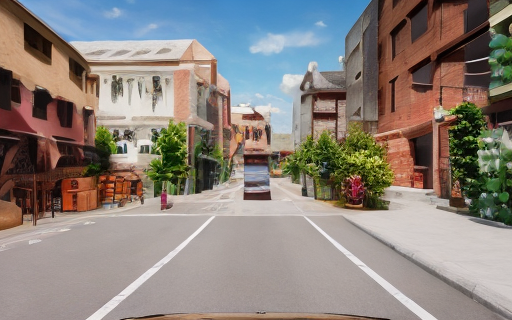}
    & \includegraphics[width=0.15\linewidth]{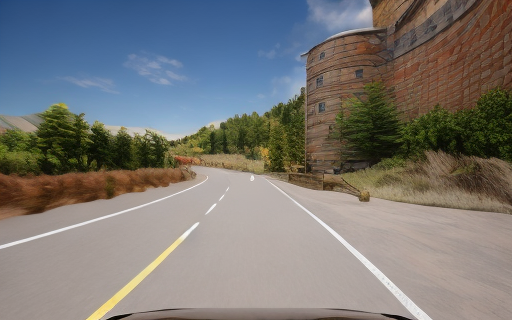}
    & \includegraphics[width=0.15\linewidth]{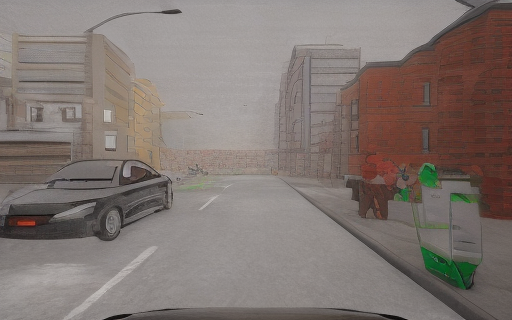}
    & \includegraphics[width=0.15\linewidth]{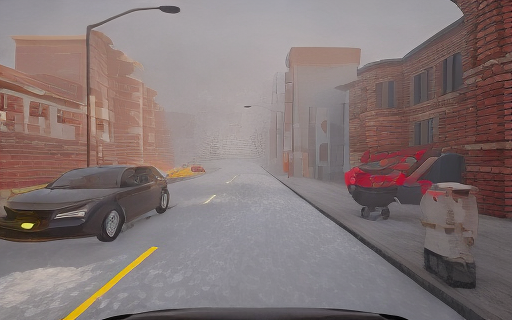}
    & \includegraphics[width=0.15\linewidth]{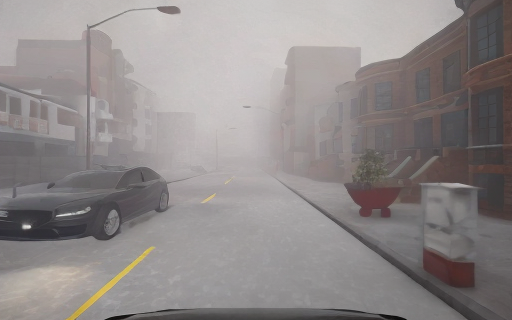}
    \\
    \multirow{-2}{*}[12mm]{\begin{subfigure}{0.2\textwidth}
        \centering
        \includegraphics[width=\linewidth]{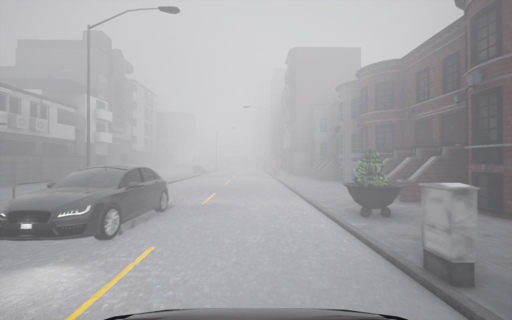}
    \end{subfigure}}
    & \rotatebox{90}{Cyclone}
    & \includegraphics[width=0.15\linewidth]{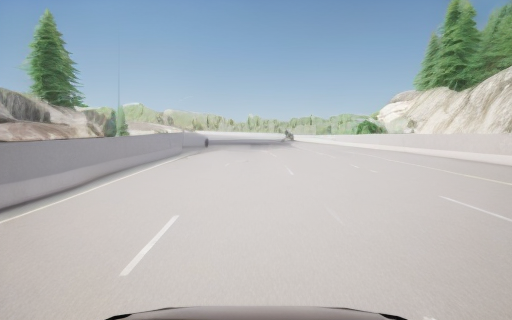}
    & \includegraphics[width=0.15\linewidth]{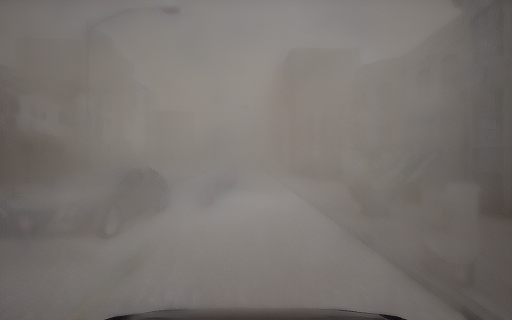}
    & \includegraphics[width=0.15\linewidth]{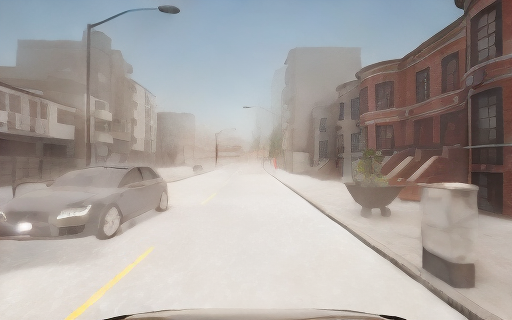}
    & \includegraphics[width=0.15\linewidth]{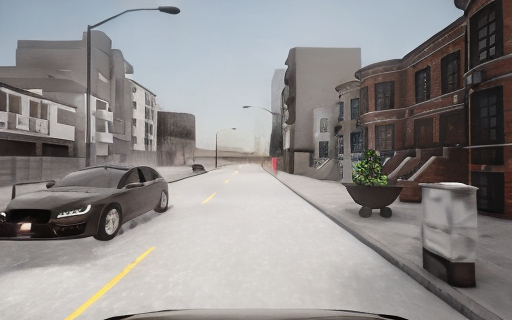}
    & \includegraphics[width=0.15\linewidth]{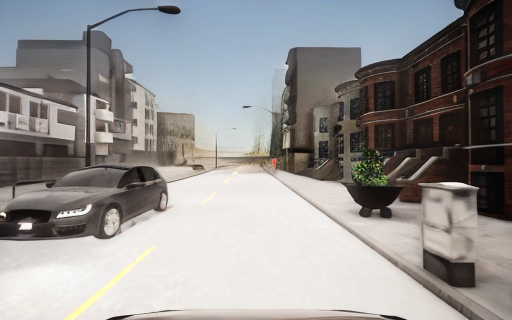}\\
    Input & {} & 100 & 1k & 10k & 20k & 50k\\
    
\end{tabular}
\caption{Comparison (foggy to clear) between CycleNet's and Cyclone's convergence over different training iterations.}
\label{fig:convergence_analysis}
\end{figure*}

We compare the convergence properties of CycleNet and Cyclone by training both models on the same data. Similar to ControlNet, CycleNet also exhibits a sudden convergence, typically occurring between 5k and 10k training iterations. However, \cref{fig:convergence_analysis} shows that CycleNet slowly overfits to the input while suffering heavily from hue color shifting. As a result, in the main paper, we re-implemented CycleNet and stopped training at around 40k iterations to avoid overfitting, whereas Cyclone can be safely trained for substantially longer. We also found that our model can quickly adapt to the conditions and produce realistic images within 15-20K iterations.

\subsection{Video Diffusion Fine-tuning}
We use Stable Video Diffusion~\cite{blattmann2023stable}  as a backbone and replace its CLIP image encoder with the text encoder. For each video sample, we encode both the input video and the corresponding ground truth video into the latent space using the VAE encoder and concatenate the noisy video latent with the input video latent.
The model is trained with a batch size of 8 and a learning rate of 3e-5 for 100k iterations.
It takes 10 GPU days for video finetuning on GPUs equivalent to NVIDIA H200.

\noindent\textbf{Temporally-consistent Generation.}
\cref{fig:temporal_consistency} showcases our model’s ability to perform video inference. We denote Cyclone and SVD-ft as our models, trained on images and fine-tuned on videos, respectively. 
While Cyclone exhibits minor color shifts and global lighting inconsistencies, the distilled video model preserves both spatial details and temporal consistency.

\def\fgsize{0.16}
\begin{figure*}
\centering
\setlength{\tabcolsep}{0.002\linewidth}
\renewcommand{\arraystretch}{1.5}
\begin{tabular}{lcccccc}
    & $t=0$ & $t=6$ & $t=12$ & $t=18$ & $t=24$ & $t=32$ \\
    \multirow{1}{*}[5mm]{\rotatebox[origin=c]{90}{Input}} & 
    \includegraphics[width=\fgsize\textwidth]{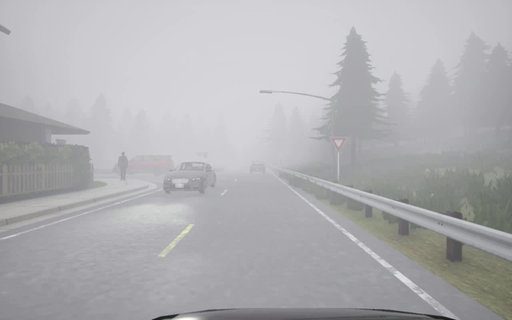} &
    \includegraphics[width=\fgsize\textwidth]{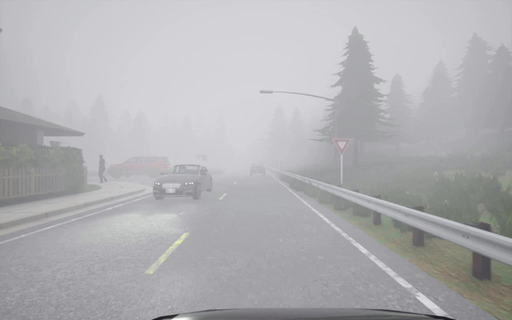} &
    \includegraphics[width=\fgsize\textwidth]{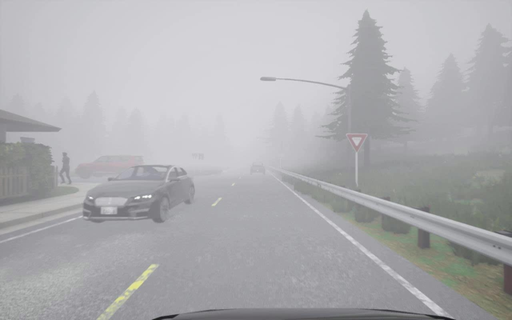} &
    \includegraphics[width=\fgsize\textwidth]{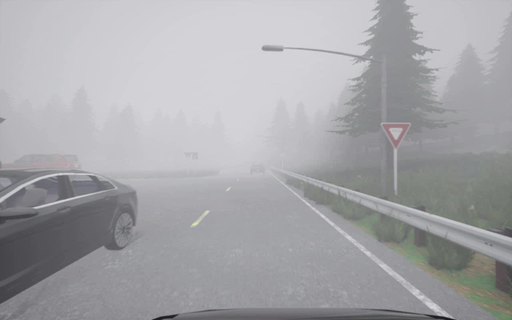} &
    \includegraphics[width=\fgsize\textwidth]{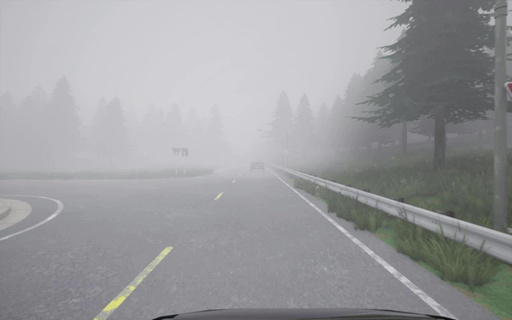} &
    \includegraphics[width=\fgsize\textwidth]{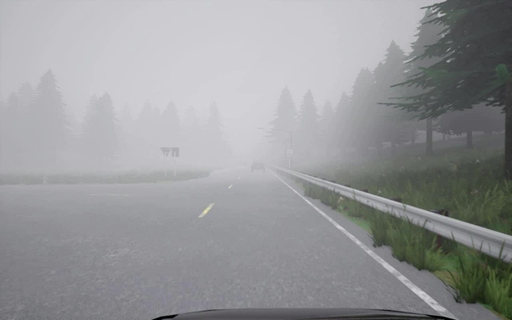}\\

    \multirow{1}{*}[7mm]{\rotatebox[origin=c]{90}{IPix2Pix}} & 
    \includegraphics[width=\fgsize\textwidth]{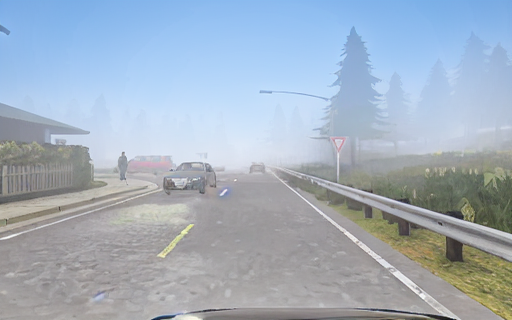} &
    \includegraphics[width=\fgsize\textwidth]{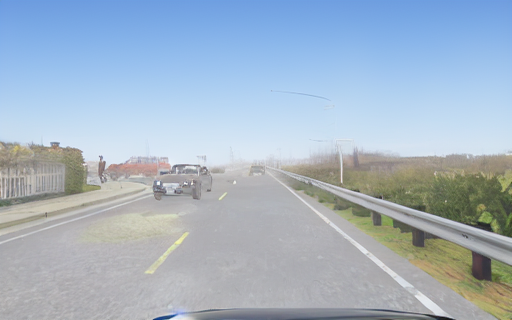} &
    \includegraphics[width=\fgsize\textwidth]{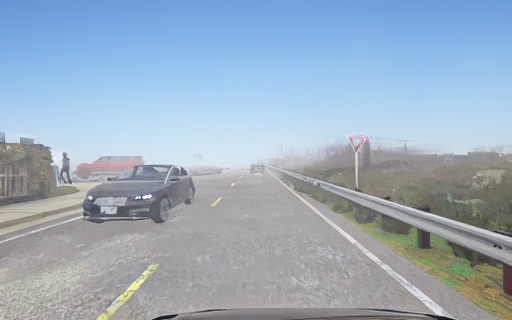} &
    \includegraphics[width=\fgsize\textwidth]{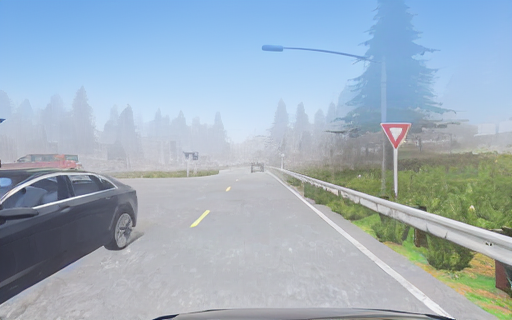} &
    \includegraphics[width=\fgsize\textwidth]{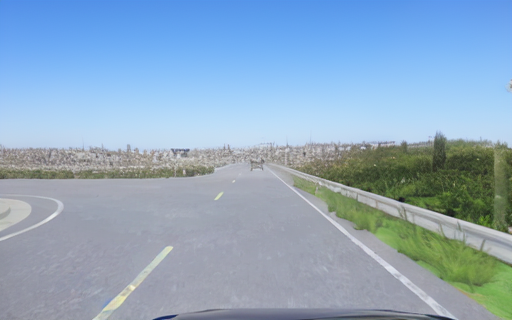} &
    \includegraphics[width=\fgsize\textwidth]{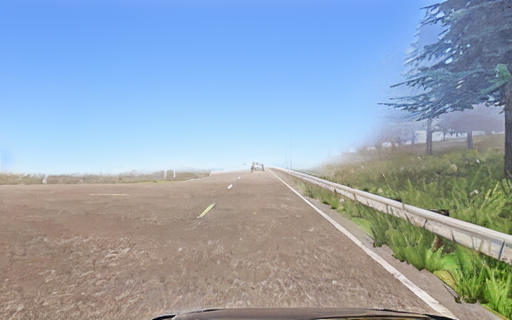}\\

    \multirow{1}{*}[7mm]{\rotatebox[origin=c]{90}{Cyclone}} & 
    \includegraphics[width=\fgsize\textwidth]{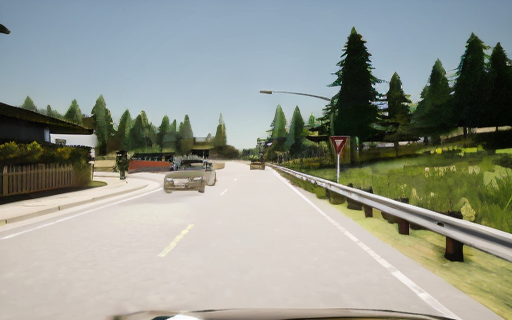} &
    \includegraphics[width=\fgsize\textwidth]{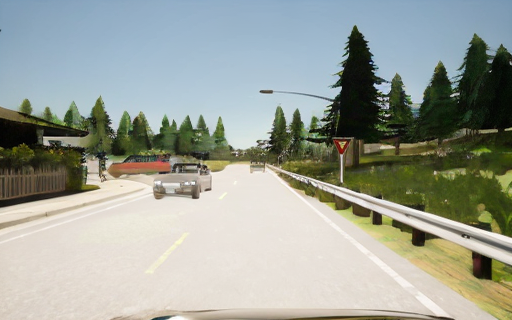} &
    \includegraphics[width=\fgsize\textwidth]{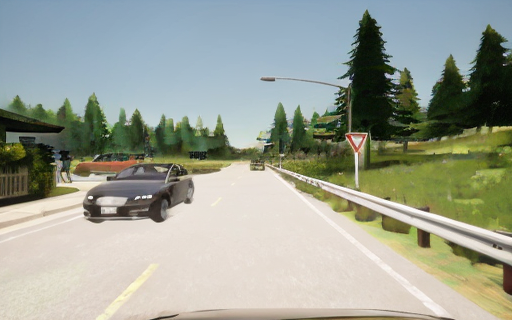} &
    \includegraphics[width=\fgsize\textwidth]{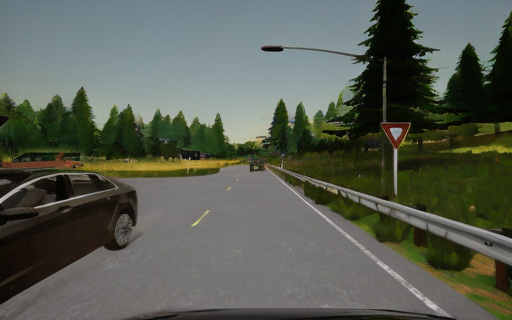} &
    \includegraphics[width=\fgsize\textwidth]{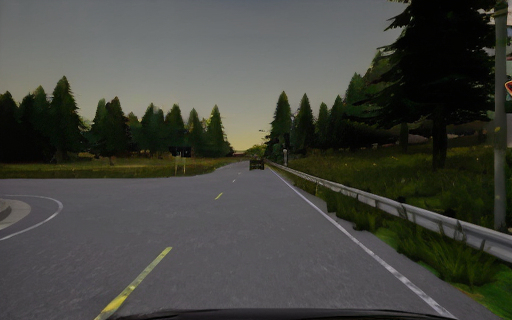} &
    \includegraphics[width=\fgsize\textwidth]{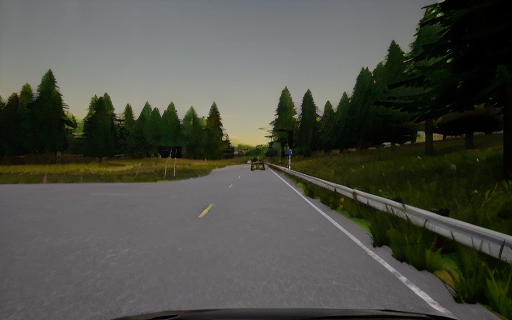}\\

    \multirow{1}{*}[7mm]{\rotatebox[origin=c]{90}{SVD-ft}} & 
    \includegraphics[width=\fgsize\textwidth]{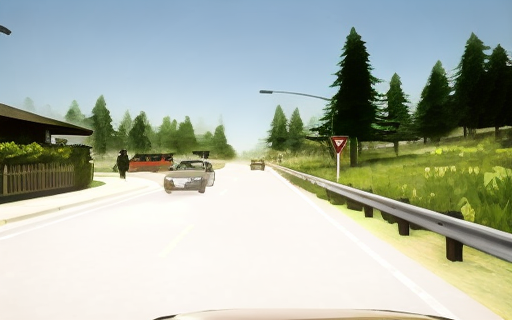} &
    \includegraphics[width=\fgsize\textwidth]{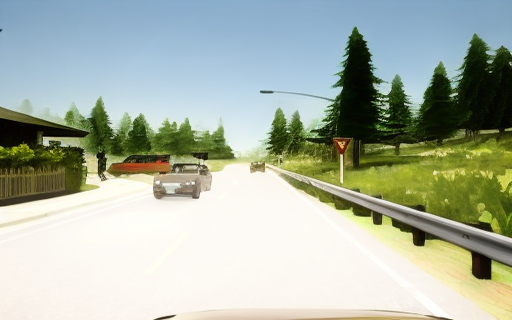} &
    \includegraphics[width=\fgsize\textwidth]{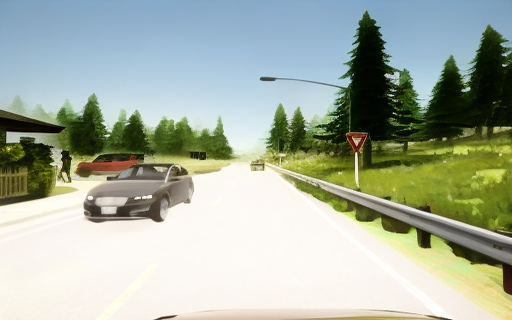} &
    \includegraphics[width=\fgsize\textwidth]{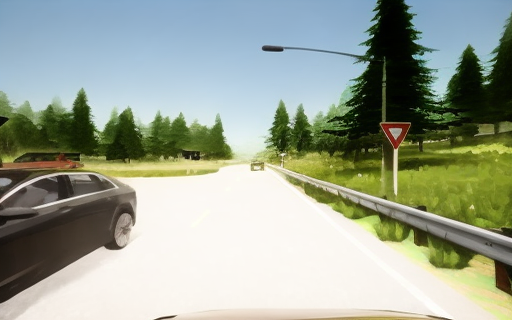} &
    \includegraphics[width=\fgsize\textwidth]{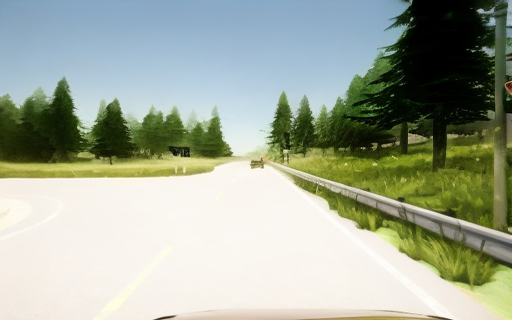} &
    \includegraphics[width=\fgsize\textwidth]{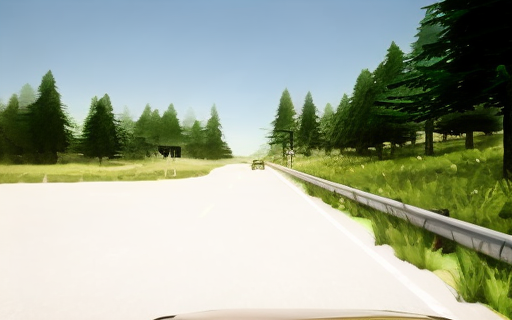}\\
\end{tabular}
\caption{Effect on temporal consistency using video diffusion model.}
\label{fig:temporal_consistency}
\end{figure*}

\section{Additional Results}
We provide additional qualitative results in \cref{fig:weather_synthesis_supmat} and \cref{fig:multi_weather_supmat}.

\def\fgsize{0.16}

\begin{figure*}
\centering
\setlength{\tabcolsep}{0.002\linewidth}
\renewcommand{\arraystretch}{1.5}
\begin{tabular}{llcccccc}
    {} & {} & $+$Foggy & $+$Rainy & $+$Snowy & $-$Foggy & $-$Rainy & $-$Snowy\\
    \multirow{2}{*}[9mm]{\rotatebox[origin=c]{90}{Input}} & {} &
    \includegraphics[width=\fgsize\textwidth]{figures/supmat/weather_synthesis/add_fog/input.png} &
    \includegraphics[width=\fgsize\textwidth]{figures/supmat/weather_synthesis/add_rain/input.png} &
    \includegraphics[width=\fgsize\textwidth]{figures/supmat/weather_synthesis/add_snow/input.png} &
    \includegraphics[width=\fgsize\textwidth]{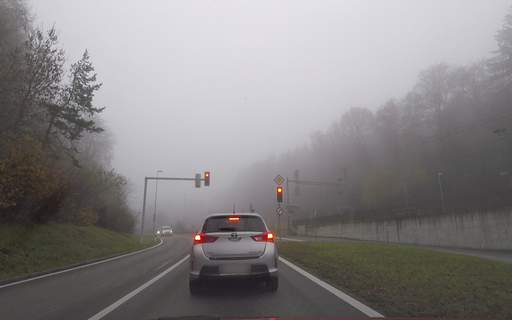} &
    \includegraphics[width=\fgsize\textwidth]{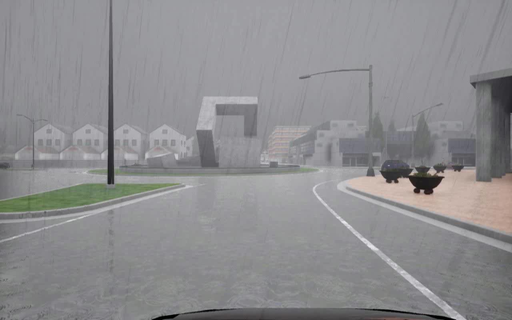} &
    \includegraphics[width=\fgsize\textwidth]{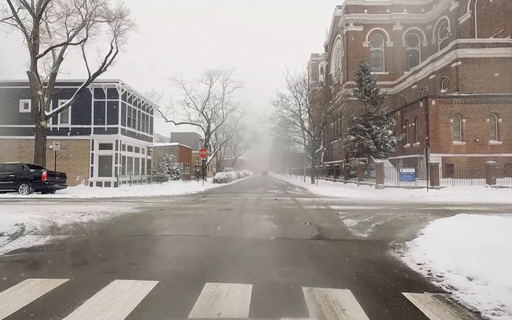}\\
    
    \multirow{2}{*}[9mm]{\rotatebox[origin=c]{90}{\small Weather}} & \multirow{2}{*}[9mm]{\rotatebox[origin=c]{90}{Diff~\cite{ozdenizci2023restoring}}} &
    & & &
    \includegraphics[width=\fgsize\textwidth]{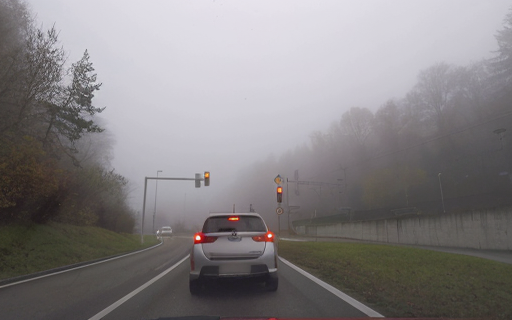} &
    \includegraphics[width=\fgsize\textwidth]{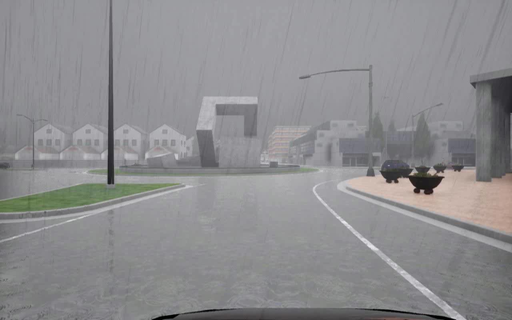} &
    \includegraphics[width=\fgsize\textwidth]{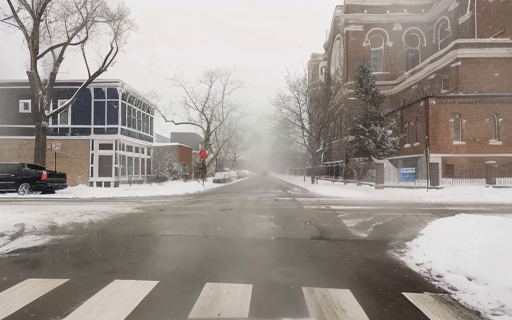}\\

    \multirow{2}{*}[9mm]{\rotatebox[origin=c]{90}{\small AwRa-}} & \multirow{2}{*}[9mm]{\rotatebox[origin=c]{90}{CLe~\cite{rajagopalan2025awracle}}} &
    & & &
    \includegraphics[width=\fgsize\textwidth]{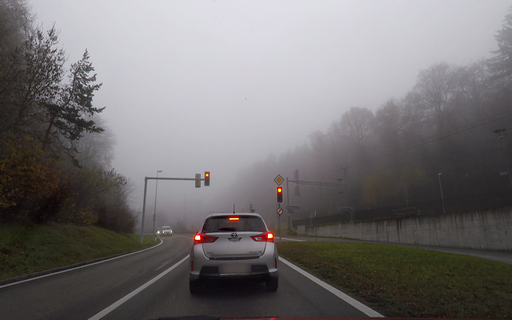} &
    \includegraphics[width=\fgsize\textwidth]{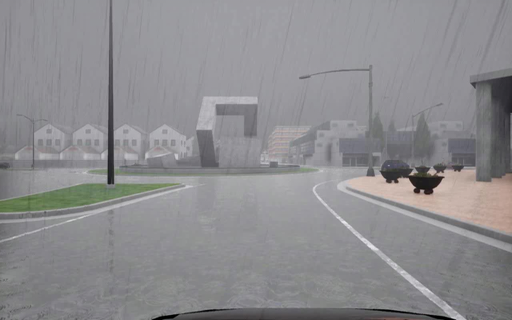} &
    \includegraphics[width=\fgsize\textwidth]{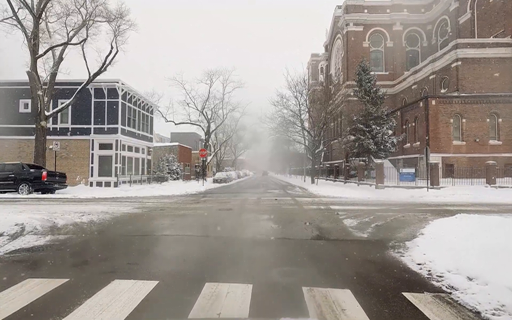}\\
    
    \multirow{2}{*}[9mm]{\rotatebox[origin=c]{90}{\small IPix2Pix}} & \multirow{2}{*}[7mm]{\rotatebox[origin=c]{90}{\cite{brooks2023instructpix2pix}}} & 
    \includegraphics[width=\fgsize\textwidth]{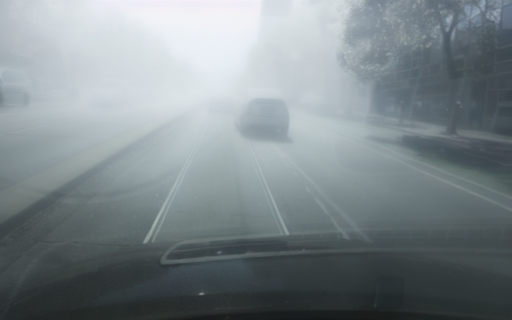} &
    \includegraphics[width=\fgsize\textwidth]{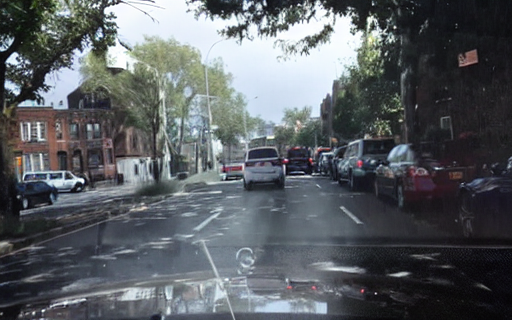} &
    \includegraphics[width=\fgsize\textwidth]{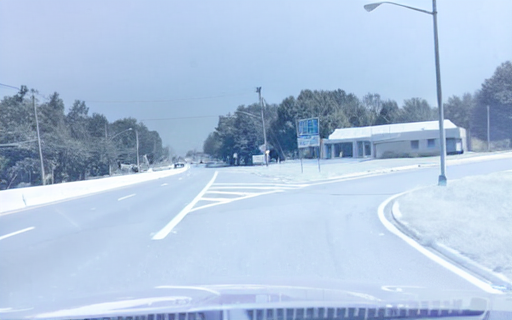} &
    \includegraphics[width=\fgsize\textwidth]{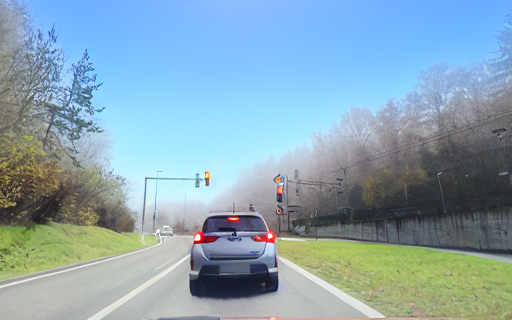} &
    \includegraphics[width=\fgsize\textwidth]{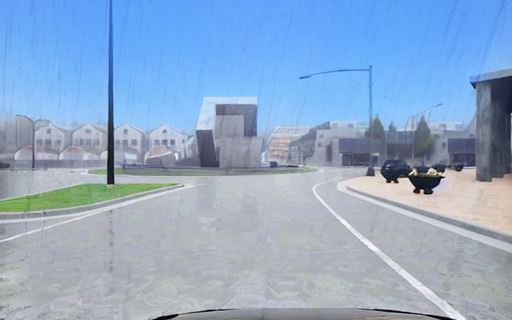} &
    \includegraphics[width=\fgsize\textwidth]{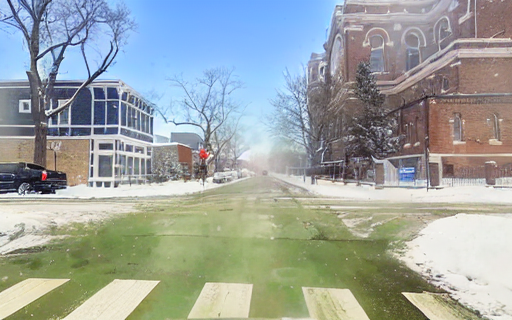}\\

    \multirow{2}{*}[9mm]{\rotatebox[origin=c]{90}{\small BAGEL}} & \multirow{2}{*}[7mm]{\rotatebox[origin=c]{90}{\cite{deng2025emerging}}} & 
    \includegraphics[width=\fgsize\textwidth]{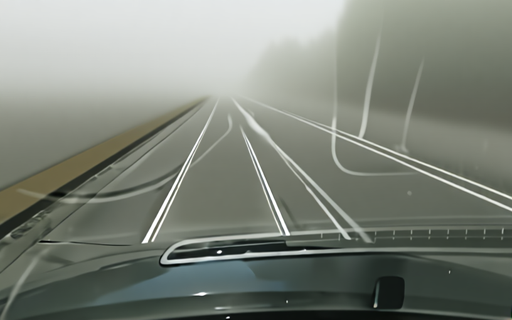} &
    \includegraphics[width=\fgsize\textwidth]{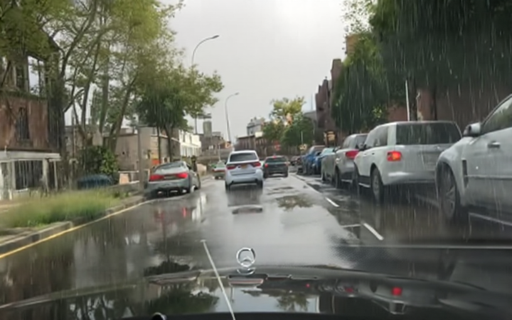} &
    \includegraphics[width=\fgsize\textwidth]{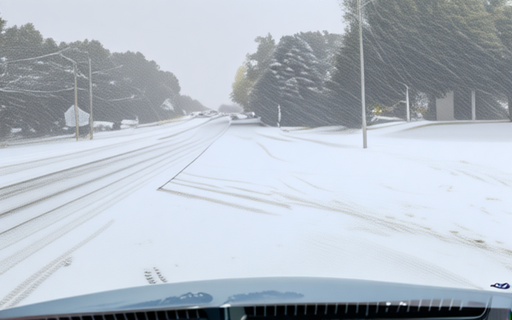} &
    \includegraphics[width=\fgsize\textwidth]{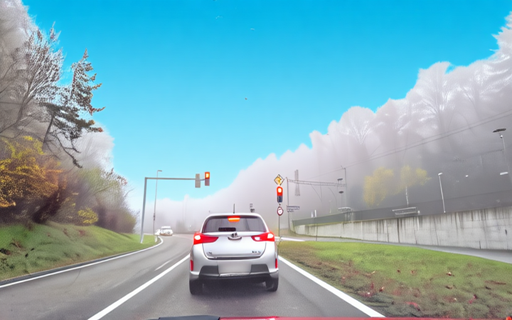} &
    \includegraphics[width=\fgsize\textwidth]{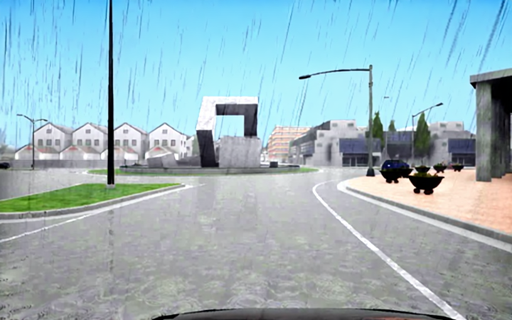} &
    \includegraphics[width=\fgsize\textwidth]{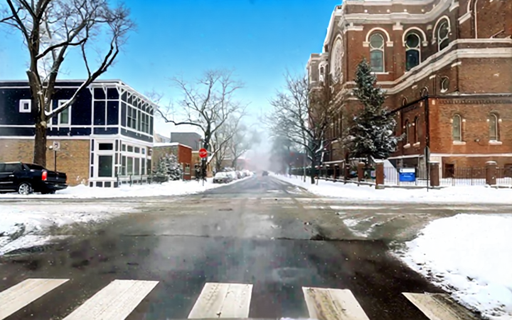}\\
    
    \multirow{2}{*}[9mm]{\rotatebox[origin=c]{90}{\small Qwen-I-E}} & \multirow{2}{*}[7mm]{\rotatebox[origin=c]{90}{\cite{bai2025qwen2}}} & 
    \includegraphics[width=\fgsize\textwidth]{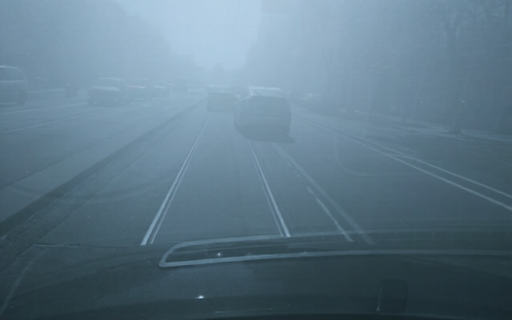} &
    \includegraphics[width=\fgsize\textwidth]{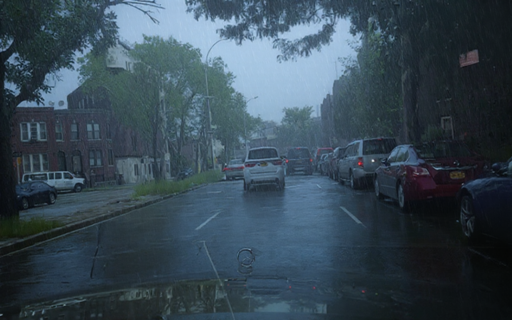} &
    \includegraphics[width=\fgsize\textwidth]{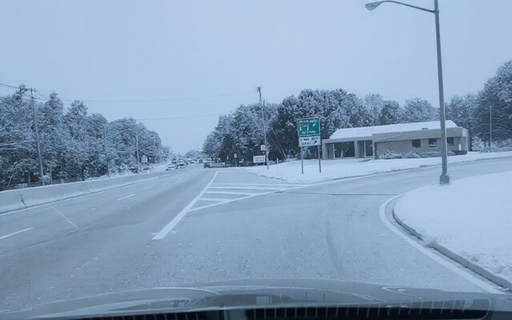} &
    \includegraphics[width=\fgsize\textwidth]{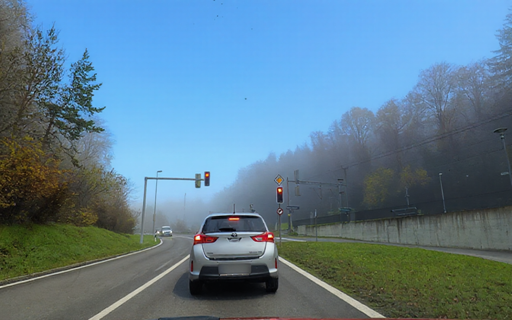} &
    \includegraphics[width=\fgsize\textwidth]{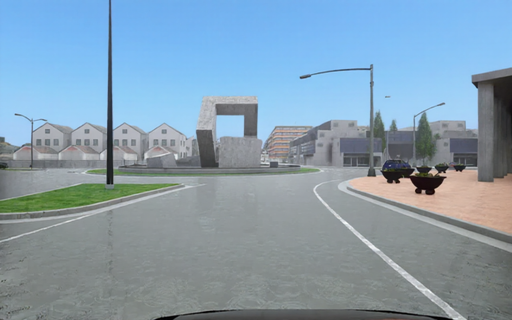} &
    \includegraphics[width=\fgsize\textwidth]{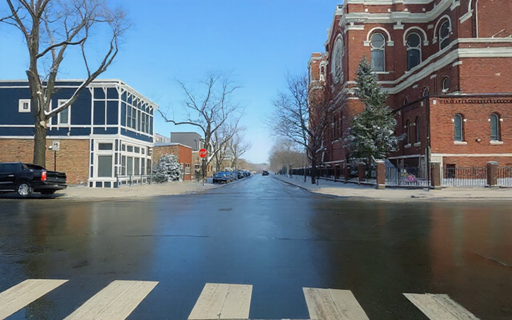}\\

    \multirow{2}{*}[9mm]{\rotatebox[origin=c]{90}{\small Tokenflow}} & \multirow{2}{*}[7mm]{\rotatebox[origin=c]{90}{\cite{geyer2023tokenflow}}} &
    \includegraphics[width=\fgsize\textwidth]{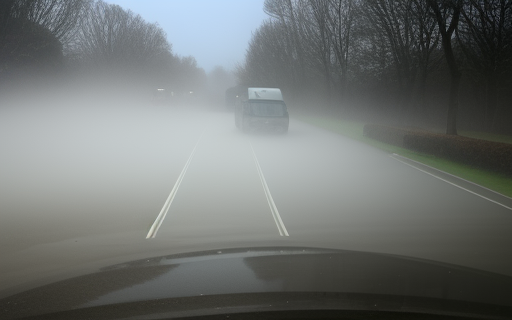} &
    \includegraphics[width=\fgsize\textwidth]{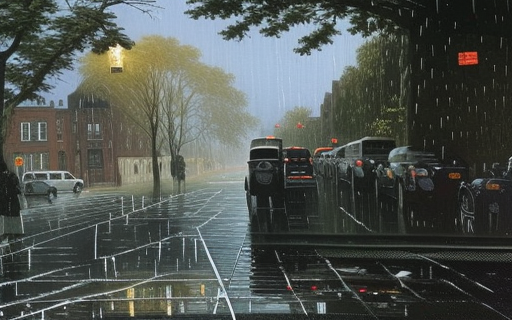} &
    \includegraphics[width=\fgsize\textwidth]{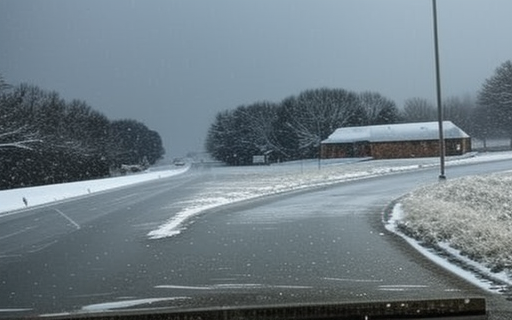} &
    \includegraphics[width=\fgsize\textwidth]{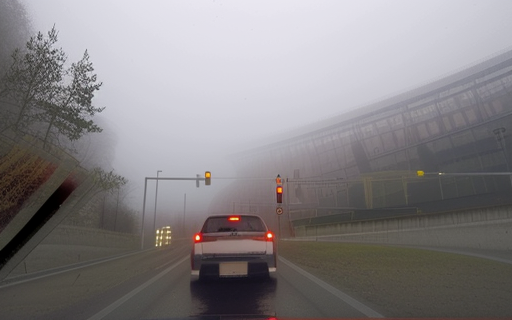} &
    \includegraphics[width=\fgsize\textwidth]{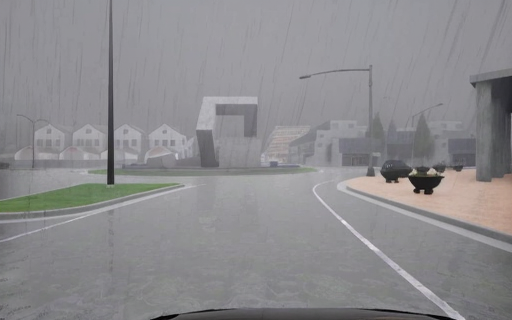} &
    \includegraphics[width=\fgsize\textwidth]{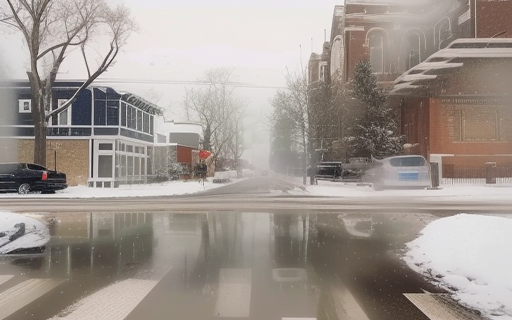}\\

    \multirow{2}{*}[7mm]{\rotatebox[origin=c]{90}{\small CycleNet}} & \multirow{2}{*}[7mm]{\rotatebox[origin=c]{90}{\cite{xu2023cyclenet}}} &
    \includegraphics[width=\fgsize\textwidth]{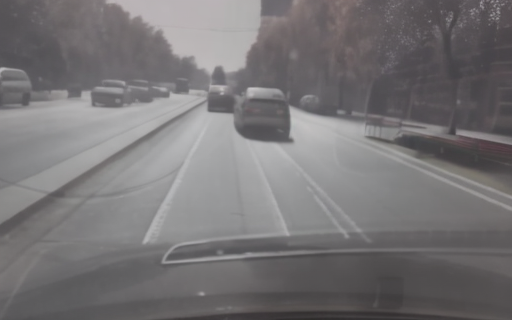} &
    \includegraphics[width=\fgsize\textwidth]{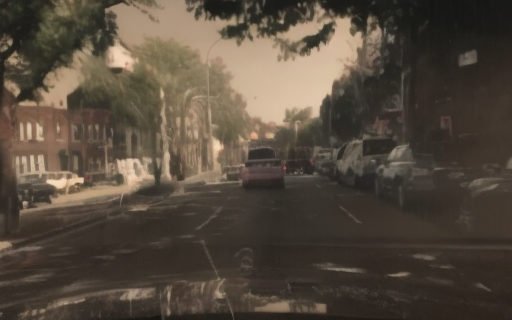} &
    \includegraphics[width=\fgsize\textwidth]{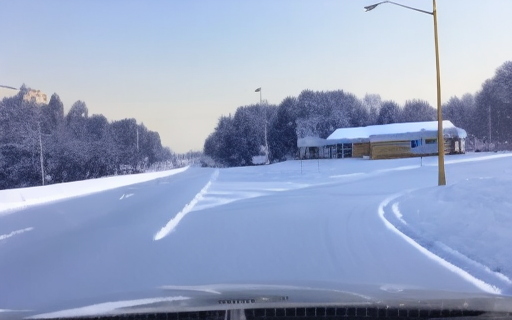} &
    \includegraphics[width=\fgsize\textwidth]{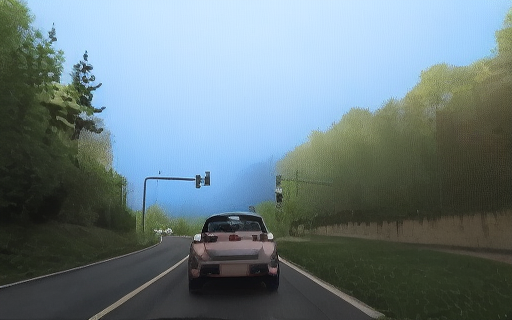} &
    \includegraphics[width=\fgsize\textwidth]{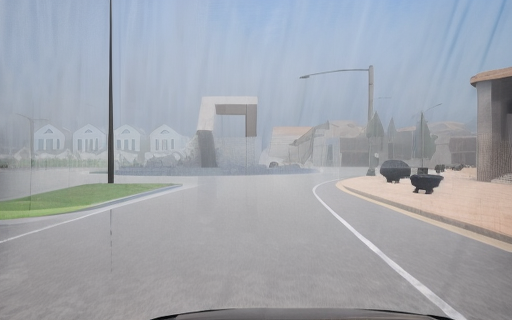} &
    \includegraphics[width=\fgsize\textwidth]{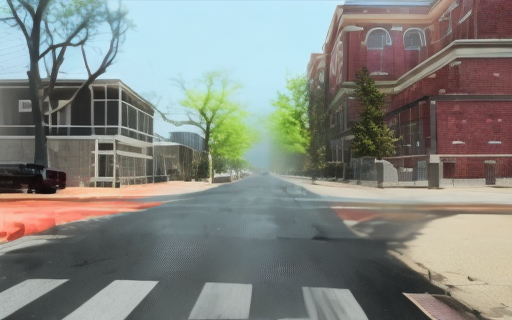}\\

    \multirow{2}{*}[7mm]{\rotatebox[origin=c]{90}{Cyclone}} & \multirow{2}{*}[8mm]{\rotatebox[origin=c]{90}{(ours)}} &
    \includegraphics[width=\fgsize\textwidth]{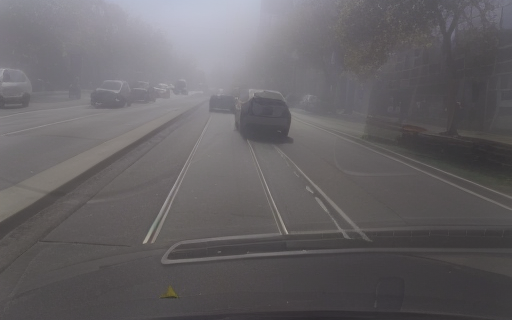} &
    \includegraphics[width=\fgsize\textwidth]{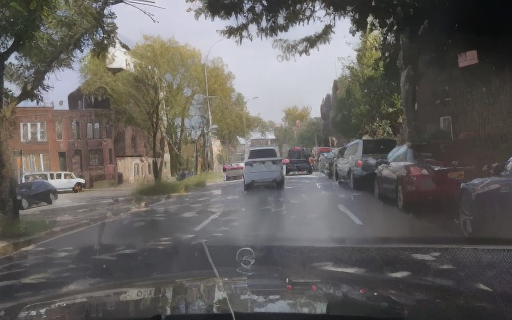} &
    \includegraphics[width=\fgsize\textwidth]{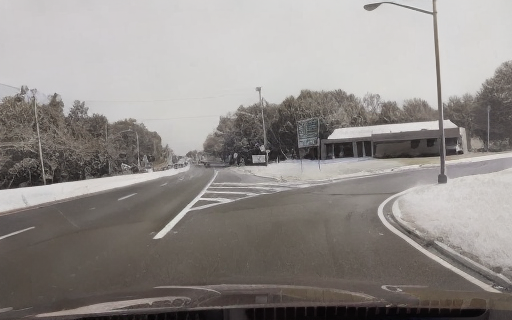} &
    \includegraphics[width=\fgsize\textwidth]{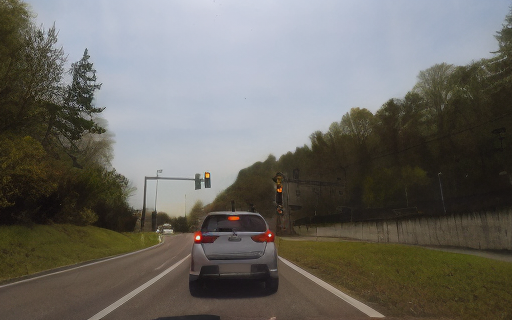} &
    \includegraphics[width=\fgsize\textwidth]{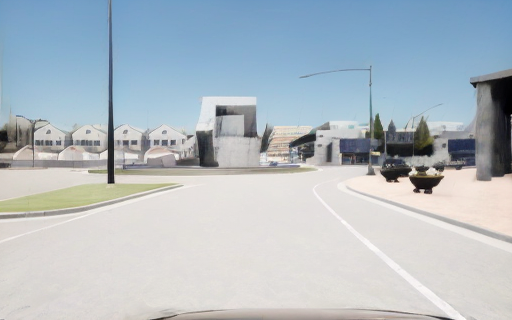} &
    \includegraphics[width=\fgsize\textwidth]{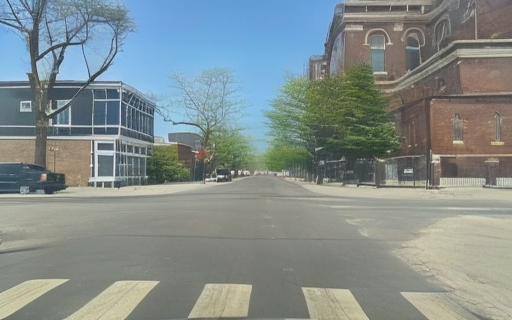}\\
    
    {} & {} & \multicolumn{3}{c}{Weather Synthesis} & \multicolumn{3}{c}{Weather Removal}\\
\end{tabular}
\caption{Additional qualitative comparison with weather editing methods.}
\label{fig:weather_synthesis_supmat}
\end{figure*} 

\begin{figure*}
\def\fgsize{0.18}
\centering
\setlength{\tabcolsep}{0.002\linewidth}
\renewcommand{\arraystretch}{0.8}
\begin{tabular}{cccccc}
    \includegraphics[width=\fgsize\textwidth]{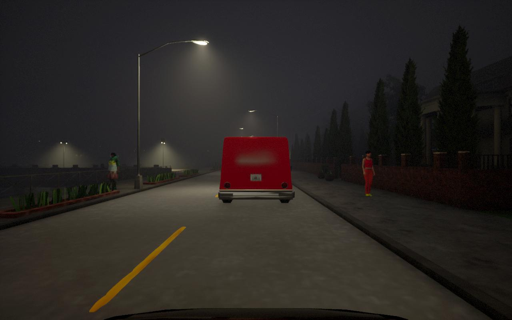} &
    \raisebox{6mm}{$\longrightarrow$} &
    \includegraphics[width=\fgsize\textwidth]{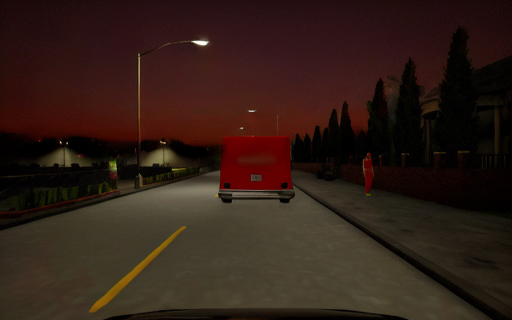} &
    \includegraphics[width=\fgsize\textwidth]{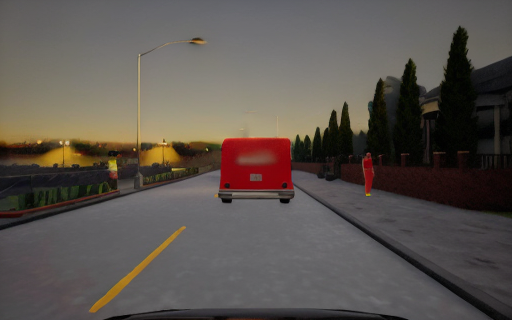} &
    \includegraphics[width=\fgsize\textwidth]{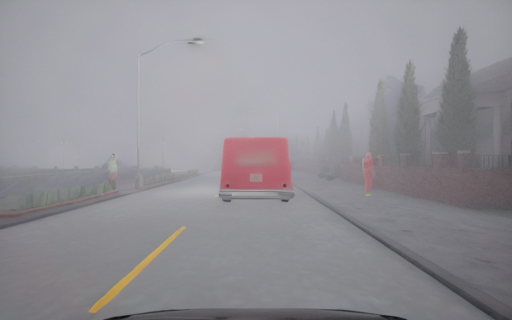} &
    \includegraphics[width=\fgsize\textwidth]{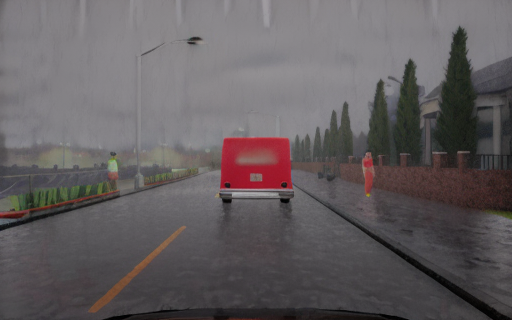} \\
    Foggy night (Input)& {} & Clear night & Clear day & Foggy day & Rainy day\\
    
    \includegraphics[width=\fgsize\textwidth]{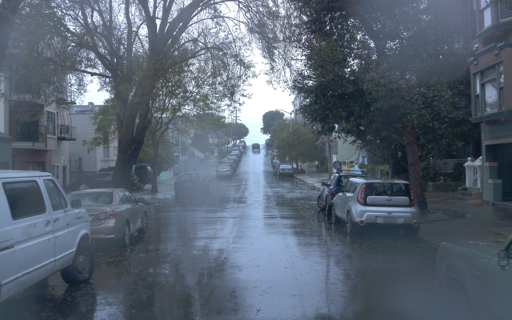} &
    \raisebox{6mm}{$\longrightarrow$} &
    \includegraphics[width=\fgsize\textwidth]{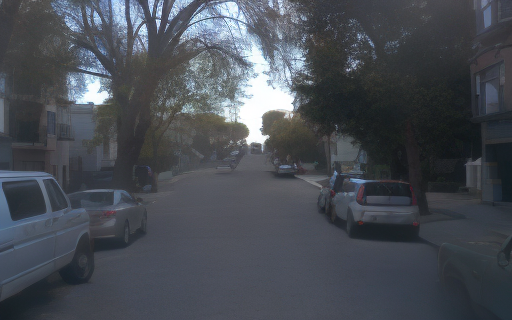} &
    \includegraphics[width=\fgsize\textwidth]{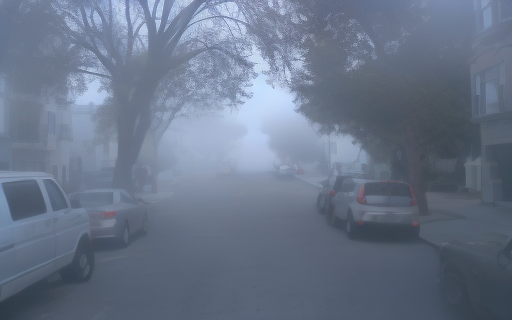} &
    \includegraphics[width=\fgsize\textwidth]{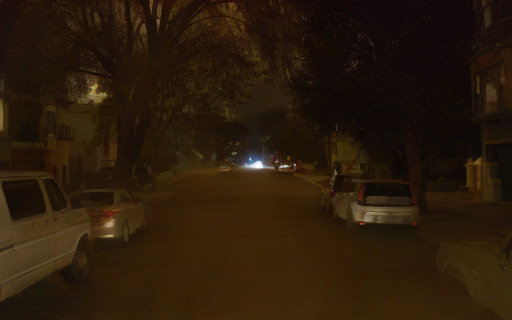} &
    \includegraphics[width=\fgsize\textwidth]{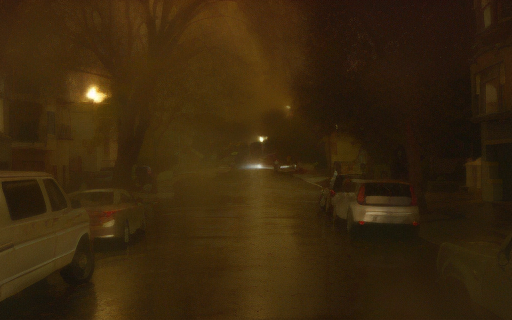}\\
    Rainy day (Input) & {} & Clear day & Foggy day & Clear night & Rainy night\\
    
    \includegraphics[width=\fgsize\textwidth]{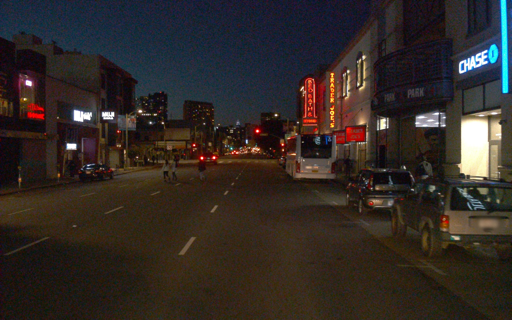} &
    \raisebox{6mm}{$\longrightarrow$} &
    \includegraphics[width=\fgsize\textwidth]{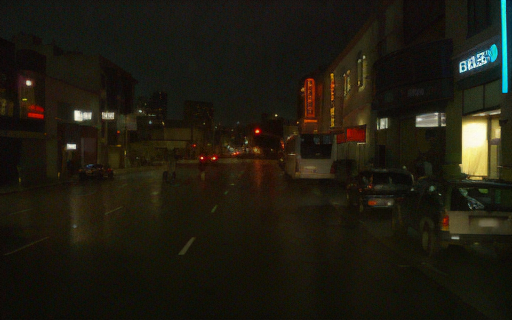} &
    \includegraphics[width=\fgsize\textwidth]{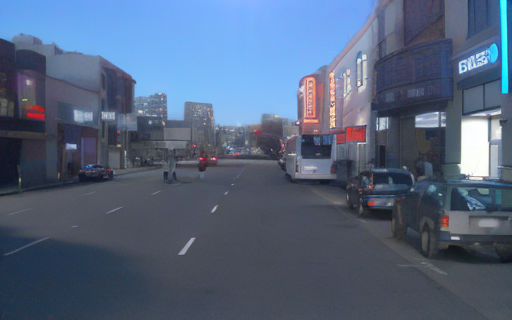} &
    \includegraphics[width=\fgsize\textwidth]{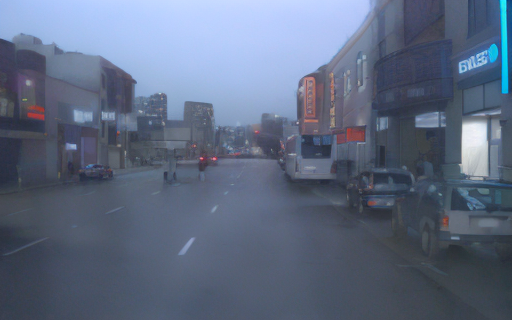} &
    \includegraphics[width=\fgsize\textwidth]{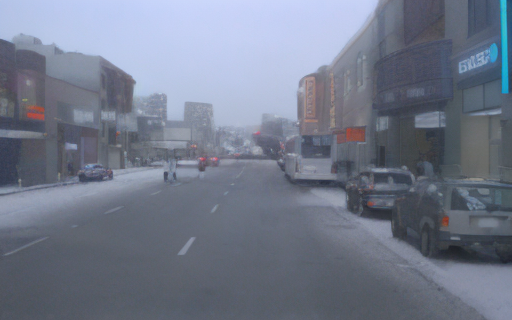} \\
    Clear night (Input) & {} & Rainy day & Clear day & Rainy day & Snowy day\\
    
    \includegraphics[width=\fgsize\textwidth]{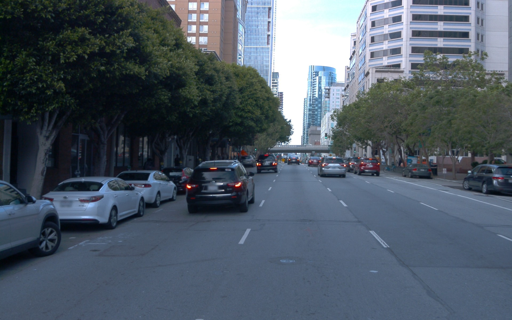} &
    \raisebox{6mm}{$\longrightarrow$} &
    \includegraphics[width=\fgsize\textwidth]{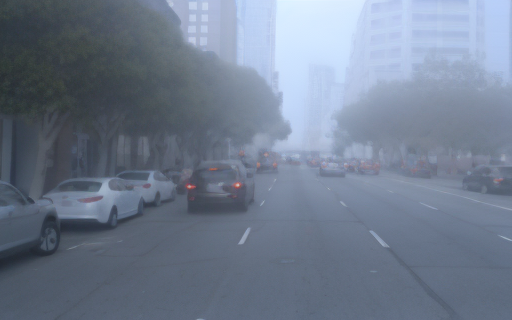} &
    \includegraphics[width=\fgsize\textwidth]{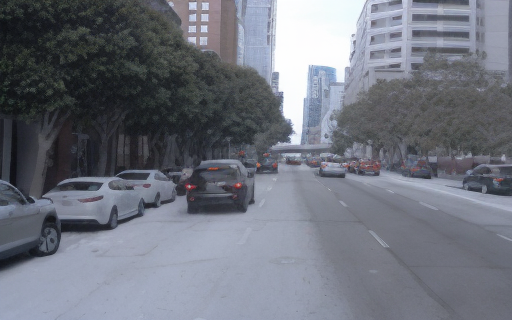} &
    \includegraphics[width=\fgsize\textwidth]{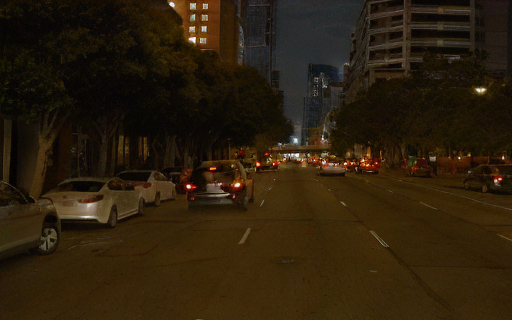} &
    \includegraphics[width=\fgsize\textwidth]{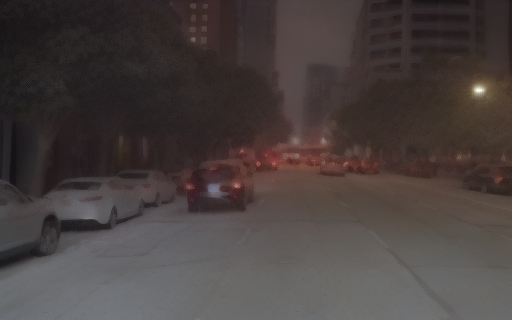} \\
    Clear day (Input)& {} & Foggy day & Snowy day & Clear night & Snowy night\\
\end{tabular}
\caption{Additional multi-weather editing showcase.}
\label{fig:multi_weather_supmat}
\end{figure*}

\subsection{Failure Cases}
\cref{fig:limitation} illustrates failure cases of Cyclone. While the model removes nighttime rain, translating from extreme low-light to daytime conditions with raindrops as occlusions makes it difficult to recover structural details. Additionally, high-frequency elements such as text are often lost due to VAE compression. A potential solution would be to adopt a more powerful VAE with better reconstruction fidelity or to apply pixel-level diffusion.

\input{figures/limitation}

\end{document}